\documentclass{article}

\usepackage[accepted]{icml2020}

\usepackage[pdftex]{graphicx}
\usepackage[utf8]{inputenc} 
\usepackage[T1]{fontenc}    
\usepackage{hyperref}       
\usepackage{url}            
\usepackage{booktabs}       
\usepackage{amsfonts}       
\usepackage{nicefrac}       
\usepackage{microtype}      
\usepackage{textcomp}
\usepackage{siunitx}
\usepackage{multirow}

\usepackage{bbm}
\usepackage{mathrsfs}
\usepackage{adjustbox}
\usepackage{makecell}
\usepackage{wrapfig}

\usepackage{amssymb,amsmath,amsfonts,bm,epsfig,graphicx,theorem,epstopdf}
\usepackage{rotating,setspace,latexsym,epsf,color,dsfont,caption,mathtools}
\usepackage{subfigure}

\usepackage{algorithm}
\usepackage{algorithmic}

\usepackage{enumitem}
\usepackage{tabularx}

\renewcommand{\algorithmicrequire}{\textbf{Input:}}

\newtheorem{theorem}{Theorem}
\newtheorem{lemma}{Lemma}
\newtheorem{assumption}{Assumption}
\newtheorem{proposition}{Proposition}
\newtheorem{corollary}{Corollary}

\newcommand{\ssf}[1]{\textrm{$\sf{#1}$}{}}

\icmltitlerunning{Learning for Dose Allocation in Adaptive Clinical Trials with Safety Constraints}

\begin{document}

\twocolumn[
\icmltitle{Learning for Dose Allocation in Adaptive Clinical Trials with Safety Constraints}



\icmlsetsymbol{equal}{*}

\begin{icmlauthorlist}

%
%

\icmlauthor{Cong Shen}{ee}
\icmlauthor{Zhiyang Wang}{to}
\icmlauthor{Sof{\'i}a S.~Villar}{goo}
\icmlauthor{Mihaela van der Schaar}{goo,ed}
\end{icmlauthorlist}

\icmlaffiliation{ee}{University of Virginia, USA}
\icmlaffiliation{to}{University of Pennsylvania, USA}
\icmlaffiliation{goo}{University of Cambridge, United Kingdom}
\icmlaffiliation{ed}{University of California, Los Angeles, USA}

\icmlcorrespondingauthor{Cong Shen}{cong@virginia.edu}

\icmlkeywords{Adaptive Clinical Trial, Multi-Armed Bandits, Dose-finding Trials}

\vskip 0.3in
]



\printAffiliationsAndNotice{}  

\begin{abstract}

Phase I dose-finding trials are increasingly challenging as the relationship between efficacy and toxicity of new compounds (or combination of them) becomes more complex. Despite this, most commonly used methods in practice focus on identifying a Maximum Tolerated Dose (MTD) by learning only from toxicity events. We present a novel adaptive clinical trial methodology, called Safe Efficacy Exploration Dose Allocation (SEEDA), that aims at maximizing the cumulative efficacies while satisfying the toxicity safety constraint with high probability.  We evaluate performance objectives that have operational meanings in practical clinical trials,  including cumulative efficacy, recommendation/allocation success probabilities, toxicity violation probability, and sample efficiency. An extended SEEDA-Plateau algorithm that is tailored for the increase-then-plateau efficacy behavior of molecularly targeted agents (MTA) is also presented. Through numerical experiments using both synthetic and real-world datasets, we show that SEEDA outperforms state-of-the-art clinical trial designs by finding the optimal dose with higher success rate and fewer patients.

\end{abstract}

\section{Introduction}


\begin{table*}[ht]
\caption{Representative adaptive clinical trial studies}
\begin{center}
{
\vspace{-0.15in}
\begin{tabular}{ c | c | c | c | c }
\hline
\textbf{Study} &  \textbf{Treatment} & \textbf{Category} & \textbf{Methodology} & \textbf{Evaluation} \\
\Xhline{3\arrayrulewidth}
\cite{Tighiouart2014} & Veliparib & CTX & EWOC-PH & simulated trial \\
\hline
\cite{Whitehead2012} & MK-0752 & CTX &  joint phase I and II design & simulated trial \\
\hline
\cite{Lee2019} & Erlotinib & MTA & extended TITE-CRM & simulated trial \\
\hline
\cite{Thiessen2010} & Lapatinib & MTA & escalation to DLT & real-world trial data\\
\hline
\end{tabular}
}
\end{center}
\vspace{-0.2in}
\label{tab:intro1}
\end{table*}


An adaptive clinical trial utilizes the accumulated results to dynamically modify its future trajectory for better efficiency and ethics, while preserving the integrity and validity of the study. Studies such as the phase I trial in Acute Myeloid Leukaemia in \cite{Yap2013} and Cancer Research UK study CR0720-11 in \cite{Whitehead2012} have suggested that even some simple forms of adaptive design lead to  better usage of resources and require fewer participants. These promising results have spawned the interest in developing adaptive clinical trial methodologies in recent years \citep{Villar2015b,Pallmann2018,Atan2019a,lee2020contextual}, which is of great importance because running an actual clinical trial on human subjects is expensive and ethically sensitive. A well-designed trial methodology with thorough theoretical and simulated investigation is widely acknowledged as a crucial first step.


Traditionally, the goal of phase I clinical trials is to identify the Maximum Tolerated Dose (MTD) of a cytotoxic (CTX) or therapeutic agent, which is then used for subsequent studies \citep{Storer1989}.  However, modern cancer phase I trials test antineoplastic agents in patients with advanced cancer stages, who have often exhausted all other available treatment options \citep{Roberts2004}. These participants usually expect therapeutic benefit from participating in the trial, which has motivated the trial design to \emph{include efficacy as a co-primary end point of phase I dose-finding studies} \citep{Yan2017,Paoletti2018}. In addition, the monotonic assumption for the dose-efficacy relationship is widely adopted in state of the art designs, which is reasonable for cytotoxic agents but may not apply to the new molecularly targeted agents (MTA) such as monoclonal antibodies (see \cite{Postel2009} for an exemplary trial that illustrates this issue). Designing adaptive clinical trials that can properly address the intrinsic conflict between learning and treatment effectiveness for general dose-response models has become an important task for phase I clinical trials. 

In addition to the well-known 3+3 design \citep{Storer1989} and continual reassessment method (CRM) \citep{OQuigley1990} (and its many variants), Bayesian approaches such as Thompson Sampling (TS) \citep{Aziz2019} and Gittins index \citep{Villar2015b,Villar2015} have been proposed in the literature for dose-finding studies. However, these methods were originally designed for simplified models that do not capture some of the unique characteristics of clinical trials, often leading to lack of randomization \citep{Villar2015}, inefficient use of side information \citep{Villar2018a}, and reduced power levels and estimation issues. Notably, for cases where the best dose for combination therapies is to be found, unknown synergistic/antagonist effects are likely to exist and naive designs will fail to identify them. For MTA, the existence of a plateau of efficacy has been discussed in \cite{Zang2014} and \cite{Riviere2017}, which indicates that the toxicity constraint must be jointly studied with the dose-efficacy relationship for certain new compounds. This is also confirmed by the real-world trial result; see \cite{Tighiouart2014}. Last but not the least, safety constraints such as minimizing the adverse events (AE) \citep{Petroni2017} have not been properly evaluated with theoretical guarantees. Table~\ref{tab:intro1} summarizes some representative studies in this direction.


In this paper, we address these challenges by developing new dose-finding methods that explicitly impose safety constraints to the allocation and recommendation of dose levels in a phase I clinical trial. Through the lens of multi-armed bandits (MAB), we propose the \textit{Safe Efficacy Exploration Dose Allocation (SEEDA)} algorithm that adaptively updates the admissible set of dose levels satisfying the safety constraints, thus limiting the exploration of doses with harmful effect. Performance analysis for SEEDA is carried out with respect to several measures that have operational meanings in clinical trials, including the probability of safety constraints violation, the average efficacy for patients, and the recommendation and allocation probabilities. Noting that SEEDA only leverages the dose-toxicity logistic model and makes no assumptions on the efficacy, we then show that, by considering the increasing-then-plateau feature of the dose-efficacy relationship for MTA, \textit{SEEDA-Plateau} leads to better performance by leveraging the unimodal structure.  Experiments on simulated datasets as well as clinical trials built from real-world datasets show that the proposed methods are capable of finding the optimal dose with higher success rate and fewer patients in most cases, compared to other state-of-the-art designs.

\section{Model and problem formulation}
\subsection{The dose-finding model}

In a phase I dose-finding clinical trial, a total of $K$ doses are given where the $k$-th dose is denoted as $d_k\in\mathcal{D}, k\in\mathcal{K}=\{1,2,...,K\}$. The performance is characterized by both \emph{efficacy} and \emph{toxicity}. We model the efficacy $X$ and toxicity $Y$ for dose $d_k$ as Bernoulli random variables with unknown probabilities $q_k$ and $p_k$, respectively, where $X=1$ ($X=0$) indicates that the dose level is effective (not effective), and $Y=1$ ($Y=0$) suggests that the dose is harmful (not harmful) to the patient\footnote{This is typically measured by the presence of absence of a dose-limiting toxicity (DLT) reported in a fixed evaluation window after administrating the drug.}.

We consider adaptive clinical trials where information learned from previous trial patients can be used in allocating doses to subsequent patients \citep{Atan2019a,Villar2015b,Aziz2019}. For the $t$-th patient, dose $I(t)$ is selected based on a policy that uses past observations, and administrated to the patient. The efficacy outcome $X_t$ and toxicity response $Y_t$ are realized based on their distributions $X_t\sim Ber(q_{I(t)})$ and $Y_t\sim Ber(p_{I(t)})$, and observed by the trialist.

We adopt a well-known dose-toxicity logistic model proposed by in  \cite{OQuigley1990} to describe the toxicity probability for different dose levels:
\begin{equation}
\label{eqn:toxicity}
    p_k(a)=\left(\frac{\tanh{d_k}+1}{2}\right)^a,
\end{equation}
where $a$ is a global parameter for all the dose levels. It can be verified that Eqn.~\eqref{eqn:toxicity} satisfies the assumption that the toxicity  monotonically increases with dose $d_k$. The unsafe dose levels are defined as those whose toxicity probabilities $p_k$'s are above a pre-determined target toxicity probability $\theta$, which is referred as the MTD threshold. Hence the toxicities of all doses can be written as $p_1\leq p_2 \leq \cdots \leq p_M < \theta < p_{M+1} \leq \cdots p_K$ where the (unknown) $M$ denotes the number of safe doses. The efficacy-dose relationship is not modeled to allow for the development of a general algorithm. The specific increase-then-plateau efficacy behavior of MTA will be exploited in Section~\ref{sec:eff}.

\subsection{Problem formulation}
\label{eqn:probform}

Several objectives are often desired for a successful dose-finding study, which are summarized as follows.
\vspace{-0.15in}
\begin{itemize}[leftmargin=*]\itemsep=0pt
\item \textbf{Successful recommendation.}  At the end of the trial ($n$ patients) a \textit{dose recommendation} $\hat{k}_n$ is made, which is desired to match the optimal dose $k^*$ that
is the lowest safe dose that achieves the highest efficacy \citep{Zang2014}: 
$k^*=\min\{k: q_k = \max_{l: l \in \mathcal{K}, p_l \leq \theta} q_l \}$.
\item \textbf{Effective treatment.} The cumulative treatment for trial participants $\sum_{i=1}^n X_t$ is desired to be maximized.  
\item \textbf{Minimal violation of the safety constraint.} There are different formulations for the safety constraint. One is to minimize 
$\mathbb{E}[\sum_{{k \in \mathcal{K}, p_k>\theta}} N_k(n)/n]$
where $N_{k}(t)$ denotes the number of times dose $k$ is allocated to the first $t$ patients. Another formulation is to minimize the probability that the average toxicity exceeds the MTD threshold.
\item \textbf{Small sample size.} Most phase I trials have a pre-determined $n$ which is decided as the minimum number of trial participants to achieve a pre-defined confidence level of successful recommendation. It is desirable to have a small $n$ for cost and efficiency considerations.
\end{itemize}
\vspace{-0.12in}
Proposing a learning model that explicitly guarantees {all} of the above objectives is elusive and non-constructive in developing the dose-allocation policy. We thus formulate dose-finding clinical trials as \textit{an online efficacy learning problem with explicit safety constraint}, and subsequently provide performance analysis on the metrics of interest. Specifically, we aim at maximizing the cumulative efficacy over a finite number of patients $n$ while simultaneously guaranteeing that the average toxicity observed from the $n$ dose allocations is kept under the probability threshold $\theta$ with high probability. This can be written as:
\vspace{-0.07in}
\begin{eqnarray}
&\text{maximize }& \mathbb{E} \left[ \sum\limits_{t=1}^n X_t \right ] \nonumber \\
&\text{subject to } & \mathbb{P}\left[\frac{1}{n}\sum\limits_{t=1}^n Y_t\leq \theta \right]\geq 1-\delta.
\label{eqn:optpro}
\end{eqnarray}
Essentially, problem formulation~\eqref{eqn:optpro} focuses on safe exploration among all the dose levels to maximize cumulative efficacies. Clinical trial designs for~\eqref{eqn:optpro} thus need to pursue both objectives of toxicity and efficacy.

\section{The SEEDA algorithm}
\label{sec:alg}

\subsection{Algorithm description}

The proposed Safe Efficacy Exploration Dose Allocation (SEEDA) design is completely described in Algorithm~\ref{alg:CDA}.  In particular, $\hat{p}_k(t)$ and $\hat{q}_k(t)$ are the estimated toxicity and efficacy, respectively, after administrating the $t$-th patient. The principle of dose selection is to first dynamically construct the {admissible set} $\mathcal{D}_1(t)$ using the Upper Confidence Bound (UCB) principle \citep{Auer2002}, where the confidence interval $\alpha(t)$ is constructed as
\vspace{-0.1in}
\begin{align}
\alpha(t)=\bar{C}_1 K\left( \frac{\log \frac{2K}{\delta}}{2t}\right)^\frac{\bar{\gamma}_1}{2},
\label{eqn:alpha}
\end{align}
where $\bar{C}_1$ and $\bar{\gamma}_1$ are algorithm parameters\footnote{See Section~\ref{appx:chooseparam} in the supplementary material for a discussion on how to select these algorithm parameters.}. Note that the admissible set consists of doses that, with high confidence, satisfy the toxicity constraint. 

Then, limiting to those in the admissible set $\mathcal{D}_1(t)$, the algorithm again applies the UCB principle (UCB-1 from  \cite{Auer2002}) to select a dose with the largest $F(p,s,n)$ for the efficacy estimate:
\vspace{-0.1in}
\begin{align}
   F(p,s,n)=p+\sqrt{\frac{c\log(n)}{s}}, \label{eqn:UCB}
\end{align}
with $c$ denoting the UCB-1 coefficient. It should be noted that \eqref{eqn:UCB} can be replaced by other UCB principles, e.g., KL-UCB \cite{Garivier2011}.


\begin{algorithm}[ht]
\caption{The Safe Efficacy Exploration Dose Allocation (SEEDA) Algorithm}
\label{alg:CDA}
\begin{algorithmic}[1]
\REQUIRE $p_k(a)$ for each $k\in\mathcal{K}$; MTD threshold $\theta$; total number of patients $n$. \\
\ENSURE $N_{k}(1)=0, \hat{p}_k(1)=0, \hat{q}_k(1)=0$, $\forall  k \in \mathcal{K}$;
Sample each dose once and set: $I(t)=t$, $\hat{q}_{I(t)}(K)=X_t$, $\hat{p}_{I(t)}(K)=Y_t$, $N_{I(t)}(K)=1$, for $t=1$ to $K$; $t=K+1$.
\WHILE{$t \leq  n$}
\STATE Compute the estimated parameter: $\hat{a}(t)=\sum_{k=1}^K w_k(t-1)\hat{a}_k(t-1)$;
\STATE Set the admissible set: $\mathcal{D}_1(t)=\{d_k\in\mathcal{D}: p_k(\hat{a}(t)+\alpha(t)) \leq \theta \}$;
\STATE Select dose: $I(t) = \arg\max_{d_k \in\mathcal{D}_1(t)} F(\hat{q}_k(t),N_k(t),t)$,\label{step:1};
\STATE Observe the revealed outcomes $X_t$ and $Y_t$;
\STATE Update estimations:
$\hat{q}_{I(t)}(t)=\frac{\hat{q}_{I(t)}(t-1)N_{I(t)}(t-1)+X_t}{N_{I(t)}(t-1)+1}$, $\hat{p}_{I(t)}(t)=\frac{\hat{p}_{I(t)}(t-1)N_{I(t)}(t-1)+Y_t}{N_{I(t)}(t-1)+1}$, $N_{I(t)}(t)=N_{I(t)}(t-1)+1$;
\vspace{-0.1in}
\STATE Update parameter estimation:
 $\hat{a}_{I(t)}(t)=\arg\min\limits_{a\in\mathcal{A}}|p_{I(t)}(a)-\hat{p}_{I(t)}(t)|$;
\STATE Update weights: $w_k(t)=N_k(t)/t$, $\forall d_k\in\mathcal{D}$;
\STATE $t=t+1$.
\ENDWHILE
\renewcommand{\algorithmicrequire}{\textbf{Output:}}
\REQUIRE $\hat{d}(n)=\arg\max_{d_k:p_k(\hat{a}(n))\leq \theta}p_k(\hat{a}(n))$.
\end{algorithmic}
\end{algorithm}

\subsection{Performance analysis}
The SEEDA algorithm is developed with the aim to solve problem~\eqref{eqn:optpro}. It is thus important to analyze (a) whether the cumulative efficacy is maximized, and (b) how often the toxicity constraint is violated. For metric (a), it can be equivalently formulated as regret minimization, i.e., the cumulative efficacy difference between the oracle policy with full information and that of the learning algorithm. Formally, the efficacy regret  is defined as
\vspace{-0.1in}
\begin{equation}
\label{eqn:regdef}
   R(n)=q^*n-\mathbb{E}\left[\sum\limits_{t=1}^n q_{I(t)}\right],
\end{equation}
where $q^*=q_{k^*}$ denotes the efficacy associated with the optimal dose defined in Section~\ref{eqn:probform},  and $a^*$ denotes the true parameter in \eqref{eqn:toxicity}.  
As for metric (b), we need to evaluate
\vspace{-0.07in}
\begin{equation*}
  e(n)=\mathbb{P}\left[\frac{1}{n}\sum\limits_{t=1}^n p_{I(t)}(a^*)> \theta \right],
\end{equation*}
in conjunction with \eqref{eqn:regdef}, i.e., whether the proposed \textit{SEEDA} algorithm minimizes $R(n)$ and satisfies $e(n)\leq \delta$ at the same time. In addition, other performance measures such as successful recommendation probability and sample efficiency are of practical interest, and we provide theoretical guarantees for them as well. Due to space limitations, all proofs are provided in the supplementary material.

\subsubsection{Cumulative efficacy}
We start the theoretical analysis by showing that for each patient $t$ in SEEDA, the dose levels whose toxicities are below the MTD threshold are included in the admissible set with high probability. This corresponds to the \emph{type I} error event that is of interest in clinical trials.
\begin{lemma}
\label{lem:1}
    $\mathbb{P}\left[ p_k(\hat{a}(t)+\alpha(t))>\theta\right]\leq\delta$, $\forall p_k(a^*)\leq \theta$.
\end{lemma}

\vspace{-0.12in}
Next we prove that with sufficient patients, the dose levels exceeding the toxicity threshold are excluded from the admissible set with high probability. This corresponds to the \emph{type II} error event in clinical trials.
\begin{lemma}
\label{lem:2}
If $t>t_1=\frac{1}{2}\left(\frac{\bar{C}_1K}{|\Delta-\epsilon|}\right)^{\frac{2}{\bar{\gamma}_1}}\log\frac{2K}{\delta}$, $\Delta=\min_{k\in\mathcal{K}}\Delta_k$, {where $\Delta_k=|a^*-p_k^{-1}(\theta)|$  represents the gap between $a^*$ and the parameter when the toxicity is at $\theta$,}
then:
\begin{gather}
\label{eqn:ti}
    \mathbb{P}\left[ p_k(\hat{a}(t)+\alpha(t))\leq \theta\right]\leq \exp(-2t\epsilon^2), \forall p_k(a^*)>\theta.
\end{gather}
\end{lemma}

Combining Lemmas~\ref{lem:1} and \ref{lem:2} leads to the main result on cumulative efficacy regret.
\begin{theorem}
\label{thm:1}
With  $t_1$ defined in Lemma~\ref{lem:2}, the regret of SEEDA can be upper bounded as:
\vspace{-0.07in}
\begin{align}
\label{eqn:regbnd}
R(n)\leq  \sum\limits_{d_k:p_k(a^*)\leq \theta}\frac{ c \log(n)}{q^*-q_k} + \left(n\delta Q+ \frac{1}{2} t_1
 +\frac{K-M}{2\epsilon^2} \right)
\end{align}
where $Q=\max_{i\in\mathcal{K}}|q_i-q_{k^*}|$ denotes the maximal single-step regret, and $\epsilon>0$ is a constant. Furthermore, if $\delta = O(\frac{1}{n})$, we have that $R(n)\leq O(\log n)$.
\end{theorem}
Theorem~\ref{thm:1} indicates that the efficacy regret is bounded by $O(\log n)$. A closer look at this scaling reveals that it consists of two parts. The first is due to the structureless model for efficacy -- we impose no assumption on the efficacy of different dose levels. The second part, which is reflected through $t_1$, is determined by the structured model for toxicity, which affects the admissible set.  As will be shown in Section~\ref{sec:eff}, with the increase-then-plateau efficacy assumption, the first $\log n$ component can be further improved.

\subsubsection{Safety constraint violation}

We now move on to analyzing the safety constraint violation. The first result is to verify whether the SEEDA algorithm indeed satisfies the safety constraint in problem~\eqref{eqn:optpro}.
\begin{theorem}
\label{thm:2}
For any given $n$, the average toxicity observed from the SEEDA algorithm satisfies
\vspace{-0.05in}
\begin{align*}
    \mathbb{P}\left[\frac{1}{n}\sum\limits_{t=1}^n p_{I(t)}-\theta\leq C_2\epsilon^{\gamma_2} \right]\geq 1-\delta,
\end{align*}
for an arbitrary $\epsilon>0$.  $C_2$ and $\gamma_2$ are problem-dependent parameters defined in Section~\ref{appx:assump} of the supplementary material. 
\end{theorem}

The safety constraint in problem~\eqref{eqn:optpro} is formulated based on the average toxicity exceeding the MTD threshold. In practice, we are often interested in minimizing the number of patients that have been exposed to unsafe dose levels, 
$\mathbb{E}[\sum_{{k \in \mathcal{K}, p_k>\theta}} N_k(n)/n]$.
Corollary~\ref{cor:penalty} analyzes this metric.

\begin{corollary}
\label{cor:penalty}
The number of unsafe dose allocations from SEEDA, i.e., the selected dose levels exceed the MTD threshold, can be bounded as:
\vspace{-0.07in}
\begin{equation*}
   \mathbb{E}\left [\sum\limits_{d_k:p_k> \theta}N_k(n) \right]  
    \leq t_1+\frac{K-M}{2\epsilon^2}.
\end{equation*}
\end{corollary}

Interestingly, Corollary~\ref{cor:penalty} indicates that unsafe dose allocations in SEEDA are upper bounded by a constant, which is linear in the number of unsafe doses $K-M$ \emph{regardless of the number of trial participants $n$}.

\subsubsection{Recommendation accuracy}

Finally, we analyze the recommendation accuracy of SEEDA at the end of the $n$-th dose allocation. 
\begin{corollary}
\label{cor:accuracy}
The probability that SEEDA recommends the {MTD} satisfies:
\begin{equation}
\label{eqn:probofoptrec}
    \mathbb{P}\left [\hat{d}(n)=\arg\max\limits_{d_k:p_k\leq \theta}p_k \right]\geq 1-\delta_1,
\end{equation}
where $\delta_1=2K\exp\left(-2\left(\frac{\Delta_{M}}{C_1K} \right)^{2\gamma_1}n \right)$.
\end{corollary}

Corollary~\ref{cor:accuracy} guarantees the finding of the {MTD} with high probability. The recommendation error rate decays \emph{exponentially} with the number of trial participants, which is a nice property.  It is worth noting that a lower bound of the minimal number of trial participants for a given accuracy requirement can be inferred from the upper bound of recommendation error rate \eqref{eqn:probofoptrec}. This is a practically important result, as sample efficiency directly relates to the cost and ethical constraints of a trial. This is further illustrated in the numerical experiments in Section~\ref{sec:sim_eff}.

\section{Extension to the increase-then-plateau efficacy model}
\label{sec:eff} 

\begin{algorithm}
\caption{The SEEDA-Plateau Algorithm}
\label{alg:CDA2}
\begin{algorithmic}[1]
\REQUIRE $p_k(a)$ for each $k\in\mathcal{K}$; MTD threshold $\theta$; total number of patients $n$. \\
\ENSURE $N_{k}(1)=0, \hat{p}_k(1)=0, \hat{q}_k(1)=0$, $\forall  k \in \mathcal{K}$; $L(1)=K$; $\eta = 2$; $l_{k} = 0$,  $\forall k \in \mathcal{K}$; Sample each dose once and set: $I(t)=t$, $\hat{q}_{I(t)}(K)=X_t$, $\hat{p}_{I(t)}(K)=Y_t$, $N_{I(t)}(K)=1$, for $t=1$ to $K$; $t=K+1$.
\WHILE{$t \leq  n$}
\STATE Compute the estimated parameter: $\hat{a}(t)=\sum_{k=1}^K w_k(t-1)\hat{a}_k(t-1)$;
\STATE Set the admissible set: $\mathcal{D}_1(t)=\{d_k\in\mathcal{D}: p_k(\hat{a}(t)+\alpha(t)) \leq \theta \}$;
\STATE Set $L(t)=\arg\max_{d_k\in\mathcal{D}_1(t)}\hat{q}_k(t)$ and increase $l_{L(t)}$ by 1; \label{alg2:1}
\STATE If $\frac{l_{L(t)}-1}{\eta+1}\in\mathbb{N}$, $I(t)=L(t)$; Otherwise $I(t)=\arg\max\limits_{ \substack{  \{L(t)-1,L(t),L(t)+1\} \\ \bigcap \mathcal{D}_1(t) }}F(\hat{q}_k(t),N_k(t),t)$;
\STATE Observe the revealed outcomes $X_t$ and $Y_t$;
\STATE Update estimations:
$\hat{q}_{I(t)}(t)=\frac{\hat{q}_{I(t)}(t-1)N_{I(t)}(t-1)+X_t}{N_{I(t)}(t-1)+1}$, $\hat{p}_{I(t)}(t)=\frac{\hat{p}_{I(t)}(t-1)N_{I(t)}(t-1)+Y_t}{N_{I(t)}(t-1)+1}$, $N_{I(t)}(t)=N_{I(t)}(t-1)+1$;
\vspace{-0.15in}
\STATE Update parameter estimation:
 $\hat{a}_{I(t)}(t)=\arg\min|p_{I(t)}(a)-\hat{p}_{I(t)}(t)|$;
\STATE Update weights: $w_k(t)=N_k(t)/t$, $\forall d_k\in\mathcal{D}$;
\STATE $t=t+1$.
\ENDWHILE
\STATE Estimate the turning point of efficacy as: 
\vspace{-0.15in}
\begin{align*}
&L_1(n)=\min\limits_{k:d_k\in\mathcal{D}_1(n)} \Big\{m\geq k: |\hat{q}_{m}(n)-\hat{q}_{m+1}(n)|     \\
&\leq \sqrt{\frac{c\log(n)}{N_m(n)}}+\sqrt{\frac{c\log(n)}{N_{m+1}(n)}}, \hat{q}_{m}(n)\leq \hat{q}_{m+1}(n) \Big\},
\end{align*}
$L_2(n)=\arg\max\limits_{d_k:p_k(\hat{a}(n))\leq \theta}p_k(\hat{a}(n)).$
\renewcommand{\algorithmicrequire}{\textbf{Output:}}
\REQUIRE  $\hat{d}(n)=\min\{L_1(n), L_2(n)\}$.
\end{algorithmic}
\end{algorithm}

The proposed SEEDA dose allocation policy is general in the sense that no efficacy model is assumed. In practice, however, efficacy often exhibits certain structure which, if utilized correctly, may further improve the performance. For conventional cytotoxic agents, efficacy monotonically increases with dose levels. The same is not true for MTAs, for which the dose-efficacy curve increases initially and then plateaus after reaching the level of saturation \citep{Zang2014,Riviere2017}. In this section, we modify the SEEDA algorithm to handle the increase-then-plateau efficacy model, and analyze its performance.

Formally, we introduce the following increase-then-plateau efficacy assumption, which holds for MTA.
\begin{assumption}
\label{ass:2}
$q_k, k\in\mathcal{K}$ satisfies $q_1\leq q_2\leq q_3 \leq \cdots \leq q_N=q_{N+1}= \cdots =q_K$.
\end{assumption}

The \textit{SEEDA-Plateau} algorithm is given in Algorithm~\ref{alg:CDA2}.  With Assumption \ref{ass:2}, the efficacy has an inherent non-decreasing structure. The key idea is to combine the selection rule of OSUB in \cite{Combes2014} and reform  step \ref{step:1} in Algorithm \ref{alg:CDA}. Note that step \ref{alg2:1} calculates $L(t)$ as the estimated dose level with the optimal efficacy and safe toxicity at $t$.  Algorithm~\ref{alg:CDA2} not only selects this dose level frequently enough, but also keeps exploring its {neighboring} dose levels.

We now analyze the regret of SEEDA-Plateau and present the result in Theorem~\ref{thm:3}. Compared to Theorem~\ref{thm:1} for SEEDA without the increase-then-plateau efficacy model, one can see that the first $\log(n)$ coefficient improves from $c \sum_{d_k:p_k(a^*)\leq \theta}(q^*-q_k)^{-1} $ to $c (q^*-q_{N-1})^{-1}$. This gain comes precisely from the increase-then-plateau efficacy model, as the unimodal structure that exploits this structure leads to $\log(n)$ regret only from the neighboring arm.

\begin{theorem}
\label{thm:3}
The regret of SEEDA-Plateau satisfies:
\begin{align} \label{eqn:thm3}
 \resizebox{0.99\linewidth}{\height}{$    
    R(n)\leq 
    \frac{c \log(n)}{q^*-q_{N-1}}+O\left(\log \log(n) \right) + \left( n\delta Q + t_1+\frac{K-M}{2\epsilon^2} \right).
 $}
\end{align}
Furthermore, if $\delta = O(\frac{1}{n})$, we have that $R(n)\leq O(\log n)$.
\end{theorem}
{The optimal dose level we have defined before can be rewritten as $k^*=\min\{M,N\}$}, the recommendation accuracy of SEEDA-Plateau is given in Theorem~\ref{thm:4}.
\vspace{-0.1in}
\begin{theorem}
\label{thm:4}
With $c$ set as $2<c<\frac{5}{2}$, the probability that SEEDA-Plateau fails to recommend the optimal dose can be bounded as:
\begin{align}
\mathbb{P}[\hat{d}(n)\neq {k^*}]\leq \frac{3}{n^c}+\delta_1.
\end{align}
\end{theorem}
Compared to Corollary \ref{cor:accuracy}, the error probability of SEEDA-Plateau is increased by $\frac{3}{n^c}$. This is due to the ambiguity of the efficacy-optimal dose and the toxicity-optimal one, which leads to the two candidate doses $L_1(n)$  and $L_2(n)$. In practice, however, this ambiguity can be eliminated via preliminary experiments.

\section{Experiments}
\label{sec:sim}

\subsection{Synthetic dataset}

\begin{table*}[h]
\caption{Recommendation \& allocation percentages of different designs. Optimal biological dose is \#3. In each cell the first row reports the mean value over 1000 repetitions, and the second row reports the (standard deviation).} 
\label{tab:set2}
\begin{center}
\vspace{-0.1in}
\adjustbox{max width=  \textwidth}{
\begin{tabular}{ l || c | c | c |c | c | c |||  c | c | c |c | c | c}
\hline
{} & \multicolumn{6}{c}{Recommended} & \multicolumn{6}{c}{Allocated} \\
\Xhline{3\arrayrulewidth}
Toxicity probabilities & 0.01 &0.05& \textbf{0.15}& 0.2 &0.45& 0.6 &
 0.01 &0.05& \textbf{0.15}& 0.2& 0.45& 0.6\\
\hline
Efficacy probabilities &0.1 &0.35& \textbf{0.6}& 0.6 &0.6& 0.6&
0.1& 0.35& \textbf{0.6} & 0.6& 0.6& 0.6\\ \Xhline{3\arrayrulewidth} 
\multirow{2}{*}{SEEDA}&0 & 1  & 47.20  & 47.40   & 4.40    & 0    & 11.18   & 9.18     &  30.76    & 31.71    & 12.06   & 5.11   \\ & (0)& (0.71)& (3.40 )&(3.41) &(2.46) &(0) &(0.58)& (1.99)& (7.76) &(7.69) & (3.45)& (0.62)\\
\hline
\multirow{2}{*}{SEEDA-Plateau} & 0.80 & 2.20  & 86.60 & 10.40&0  &0  & 7.83  & 8.98  & 30.12  & 37.17  &14.91 & 1.00 \\ &(0.32) &(1.96) &(8.58) &(3.65) &(0) &(0) &(1.61)&(4.21)&(6.01)&(7.54) &(3.02)&(0.61)\\
\hline
\multirow{2}{*}{Independent TS} & 2.60 & 9.40  & 44.60 & 35.40& 6.60 & 1.40&     3.66 & 7.26 & 22.22  & 21.00  & 22.26  & 23.60  \\ &(2.47) &(3.86)&(10.25)&(10.32)&(2.96)&(0.69)&(0.97)&(3.85)&(15.47)&(10.44)&(10.43)&(9.22)\\
\hline
\multirow{2}{*}{KL-UCB}  &0.20& 4.60& 48.80& 43.60& 2.80& 0&       10.93& 7.16& 21.33& 20.91& 21.21& 18.46\\ &(0.13) &(2.71)&(11.68) &(11.36) &(2.64)&(0)&(0.81)&(0.94)&(10.52) &(11.31) &(10.92) &(11.10) \\
\hline
\multirow{2}{*}{UCB} &0& 2.40& 54.00& 40.40& 3.20& 0&        5.45& 9.50& 22.13& 20.93& 20.43& 24.57
\\ &(0) &(2.15) &(9.92) &(9.05) &(3.09) &(0) &(0.49) &(1.16) &(2.11) &(2.24) &(2.15) &(2.11)\\
\hline
\multirow{2}{*}{3+3} & 0& 2.40& 12.00& 17.60& 45.20& 22.80&     16.04& 17.82& 20.19& 18.12 & 16.81& 5.82\\ &(0) &(0.41)&(4.31)&(5.31)&(7.15)&(4.35)&(5.12)&(4.23)&(10.25)&(9.15)&(8.15)&(4.12)
\\
\hline
\multirow{2}{*}{CRM} & 0& 0& 0& 33.80 & 65.80& 0.40&        0.12& 0.35& 2.62& 33.90 & 57.69& 5.33 \\&(0)&(0)&(0)&(8.26)&(10.63)&(0.40)&(0.11)&(0.25)&(0.32)&(10.21)&(11.24)&(0.23)
\\
\hline
\multirow{2}{*}{MCRM} &0& 0& 0.20& 61.00 &38.80& 0&  1.47 &1.18& 5.64& 55.48& 34.63&,1.60 \\&(0) &(0) &(0.15) &(9.67) &(8.65) &(0)&(0.24) &(0.67) &(3.62) &(8.63) &(7.65) &(0.67)
\\
\hline
\multirow{2}{*}{Multi-obj} & 0.81& 3.23& 47.90& 41.03& 5.88& 1.15&      18.42& 21.92& 23.36& 18.48& 9.92& 7.89 \\&(0.19) &(0.94) &(11.86) &(11.90) &(1.80) &(0.36) &(6.07) &(5.55) &(6.68) &(7.05) &(5.17) &(4.65)
\\
\hline
\end{tabular}
}
\end{center}
\vspace{-0.1in}
\end{table*}

To investigate the operational characteristics and evaluate the performance of the proposed adaptive designs, we present an experimental study with $K = 6$ dose levels and $n=300$ trial cohorts, with each cohort consists of 3 patients.  The estimation is updated after observing all individual outcomes from a cohort. All experiment results are obtained with 1000 trial repetitions.  The MTD threshold is set as $\theta=0.35$. 

The trial setup is the same as \cite{Riviere2017} and \cite{Zang2014}, and we have simulated eight different efficacy and toxicity scenarios\footnote{We remark that although no real-world trial data is utilized in the experiment, this approach is commonly accepted in clinical trials as the first-step study for a new methodology; see \cite{Whitehead2012,Yap2013,Zang2014,Riviere2017}.}.  Due to the space limitation, we only report the results of the first scenario, where efficacy reaching the maximal value (the optimal dose) before toxicity hits MTD threshold. Additional results for this setting as well as the other seven scenarios are reported in Section~\ref{appx:sim:setting2} to \ref{appx:sim:5setting} in the supplementary material.   


The following baseline designs are used for comparison (whenever appropriate), whose details can be found in the supplementary material: 3+3, CRM, MCRM, Independent TS, KL-UCB, UCB-1, and multi-objective bandits.  Note that MTA-RA and other TS variants in \cite{Riviere2017} are not included because they assume a different truncated efficacy model, which needs to be perfectly known to the algorithm. For algorithms that require prior information of toxicity and efficacy, they are set as $[0.02, 0.06, 0.12, 0.20, 0.30, 0.40]$ and $[0.12, 0.20, 0.30, 0.40, 0.50, 0.59]$, respectively.  

\subsubsection{Recommendation and allocation accuracy}

We  report the allocation and recommendation percentages of each dose for all considered designs in Table~\ref{tab:set2}.  Dose 3 (in bold font)  is the optimal biological dose  for this scenario. However, we comment that dose 4 also satisfies the optimality condition without violating the safety constraint. Nevertheless, it has a higher toxicity probability (although still below MTD) without increasing efficacy; thus less preferable to Dose 3.  We note that for all the considered designs, the recommendation rule is $\hat{d}(n)=\arg\max_{k:\hat{p}_k(n)\leq \theta}\hat{q}_k(n)$, where $\hat{q}_k(n)$ and $\hat{p}_k(n)$ are the final estimations of toxicity and efficacy for dose level $d_k$, respectively. This suggests that safety constraint is considered in recommendation.

\begin{figure}[h]
\centerline{
	\subfigure{
        { \includegraphics[width=0.24\textwidth]{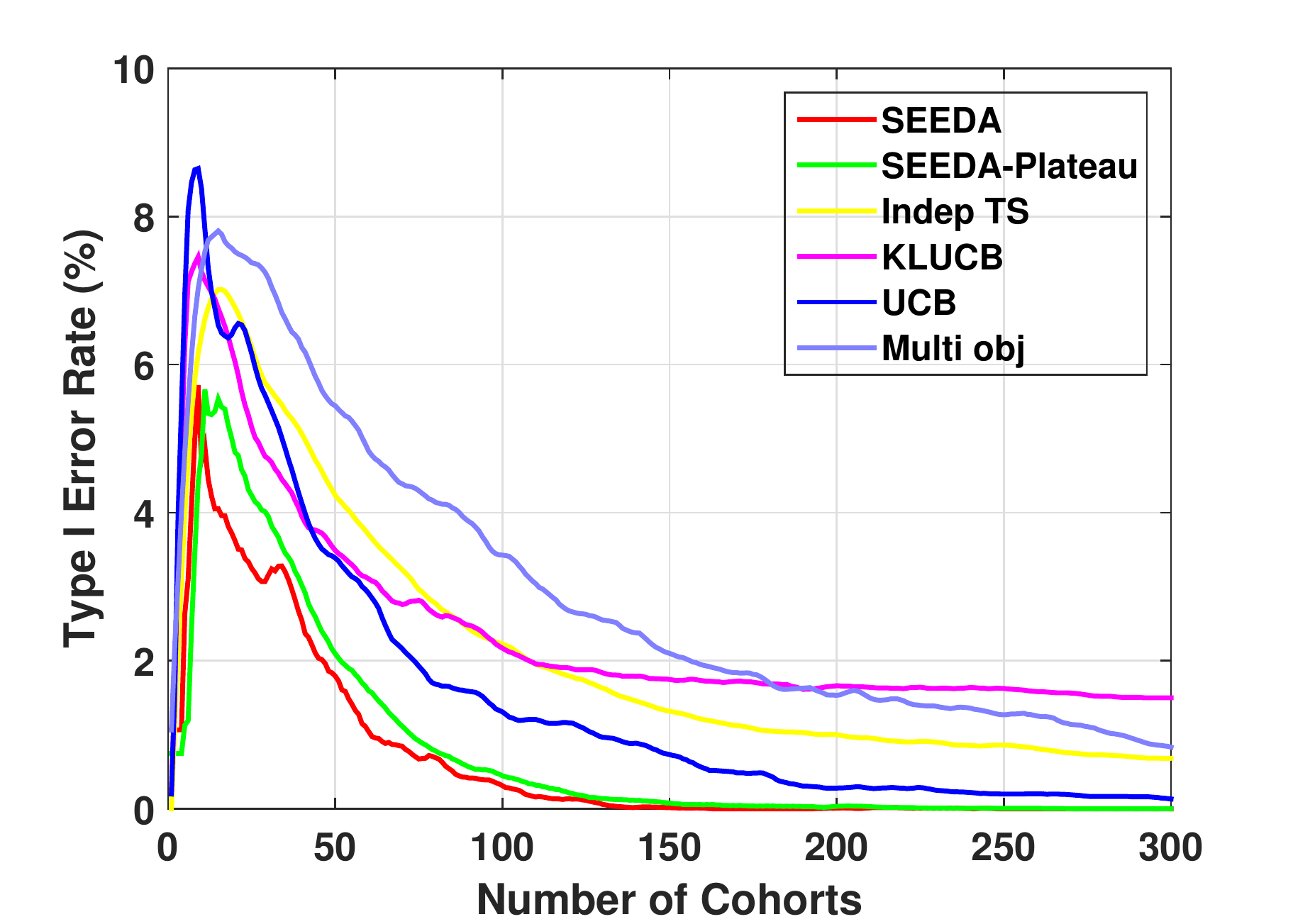}} }
        \hfil
        \subfigure{
        { \includegraphics[width=0.24\textwidth]{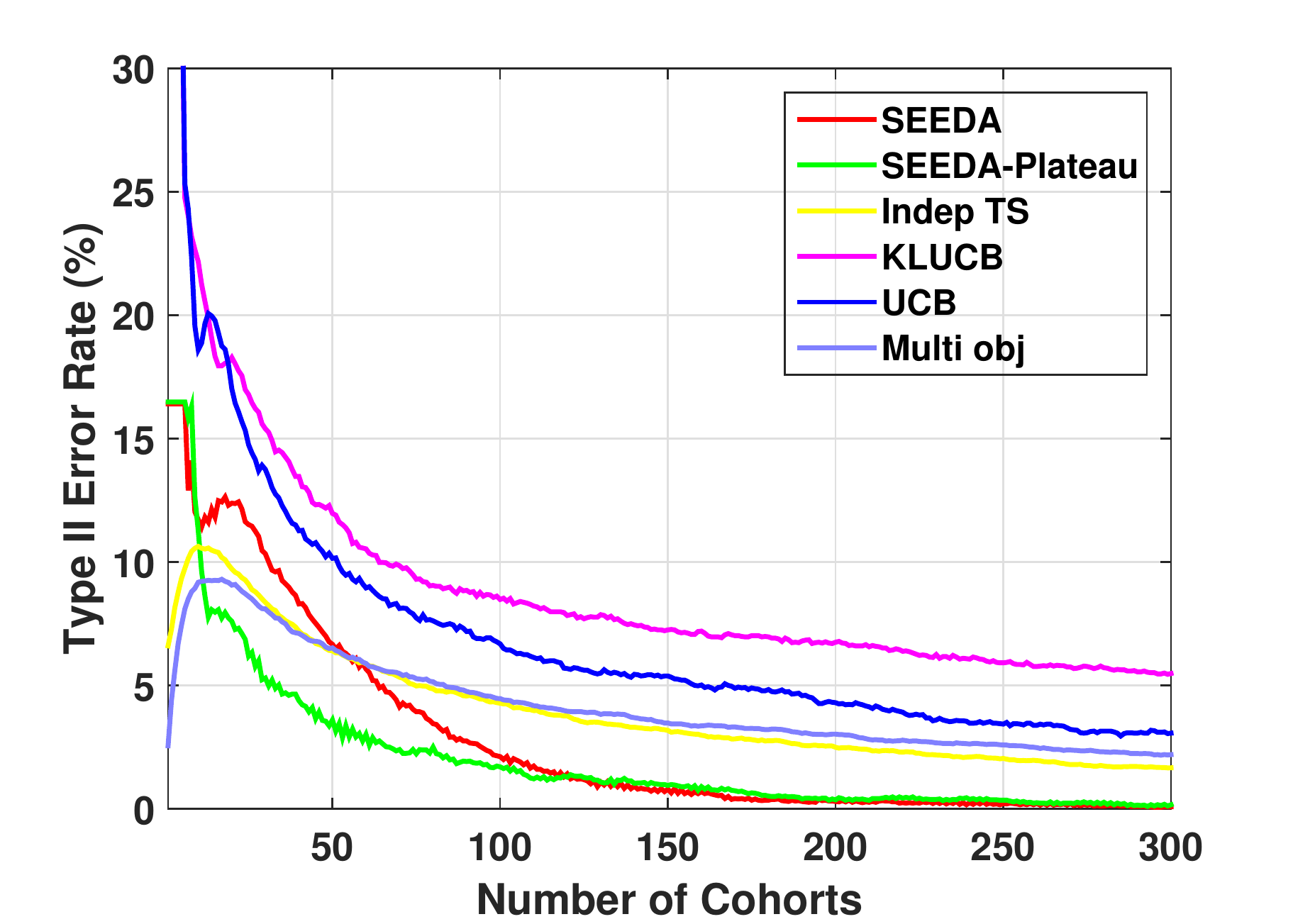}}  }
        }
    \caption{Type I (left) and type II (right) error rates as a function of number of cohorts.}
    \label{fig:type-err-set2}
\end{figure}

We can see from the results that SEEDA almost equally recommends dose 3 and 4 with a total probability of 94.6\%. This is because the algorithm cares about maximizing efficacy without violating safety constraint, and both dose 3 and 4 satisfy such conditions. As a result, SEEDA treats both equally as the optimal solution. However, by leveraging the increase-then-plateau model assumption, SEEDA-Plateau can further break the ``tie'' between dose 3 and 4, and correctly recognize that dose 3 is the optimal biological dose: it chooses dose 3 at 86.6\% while dose 4 only 10.4\%. We see that the gain of SEEDA-Plateau is significant over all the other designs (even compared to SEEDA). For a more detailed understanding of the recommendation accuracy, the corresponding type I and type II error rates (definitions are given in Section~\ref{appx:sim:baseline} in the supplementary material) are plotted in Fig.~\ref{fig:type-err-set2}, and we observe that both SEEDA and SEEDA-Plateau outperform other baseline methods over the range of cohorts.

As for allocation, we observe that both SEEDA and SEEDA-Plateau concentrate at dose 3 and 4, while spending very little budget on both tail ends of the dosage. In particular, SEEDA-Plateau  allocates the fewest percentages (1\%) of patients to the most toxic dose 6 among all designs.  


\subsubsection{Convergence and safety violation}

\begin{figure}[h]
\centerline{
        \subfigure{
        { \includegraphics[width=0.24\textwidth]{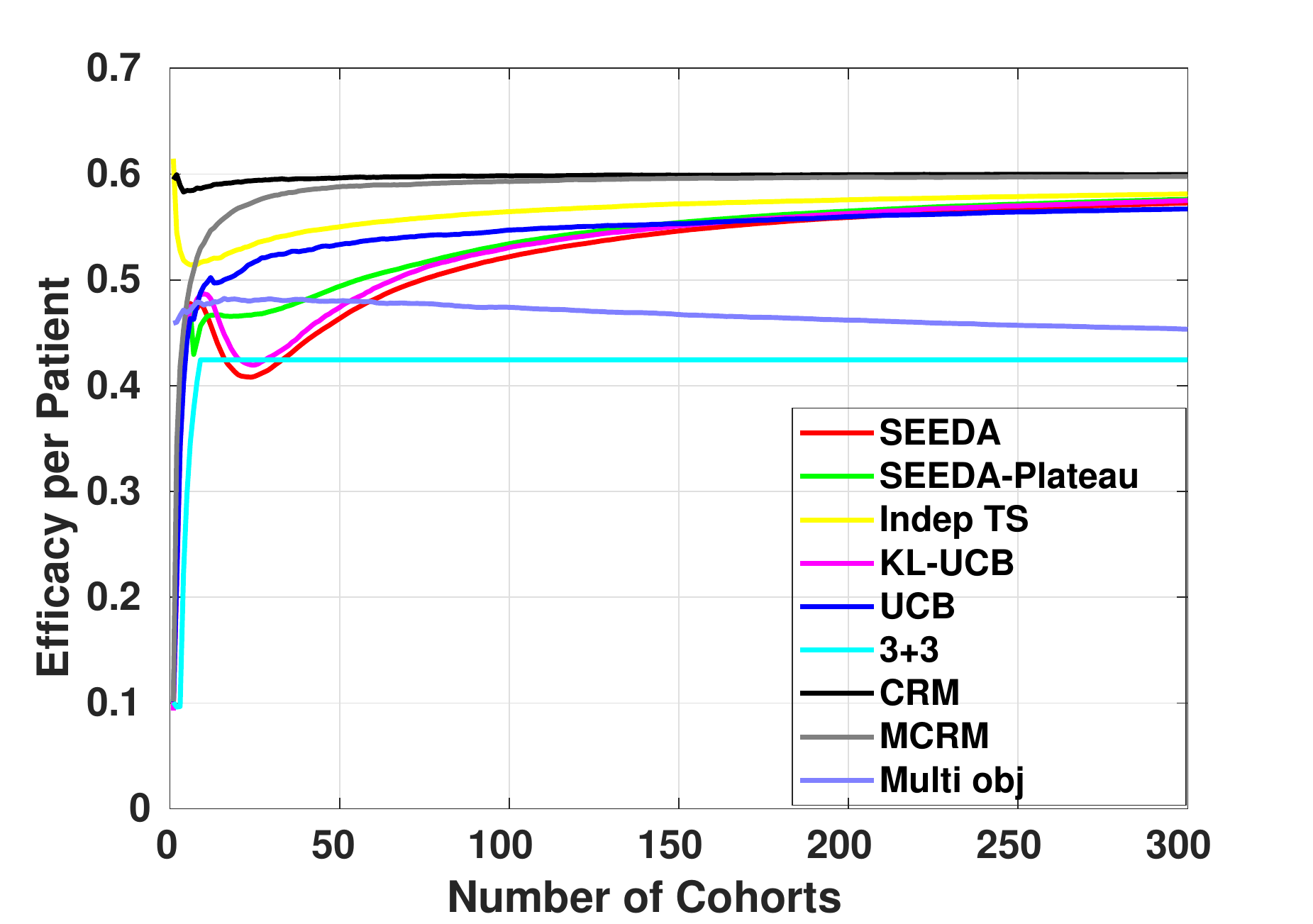}} }
        \hfil
       \subfigure{
        { \includegraphics[width=0.24\textwidth]{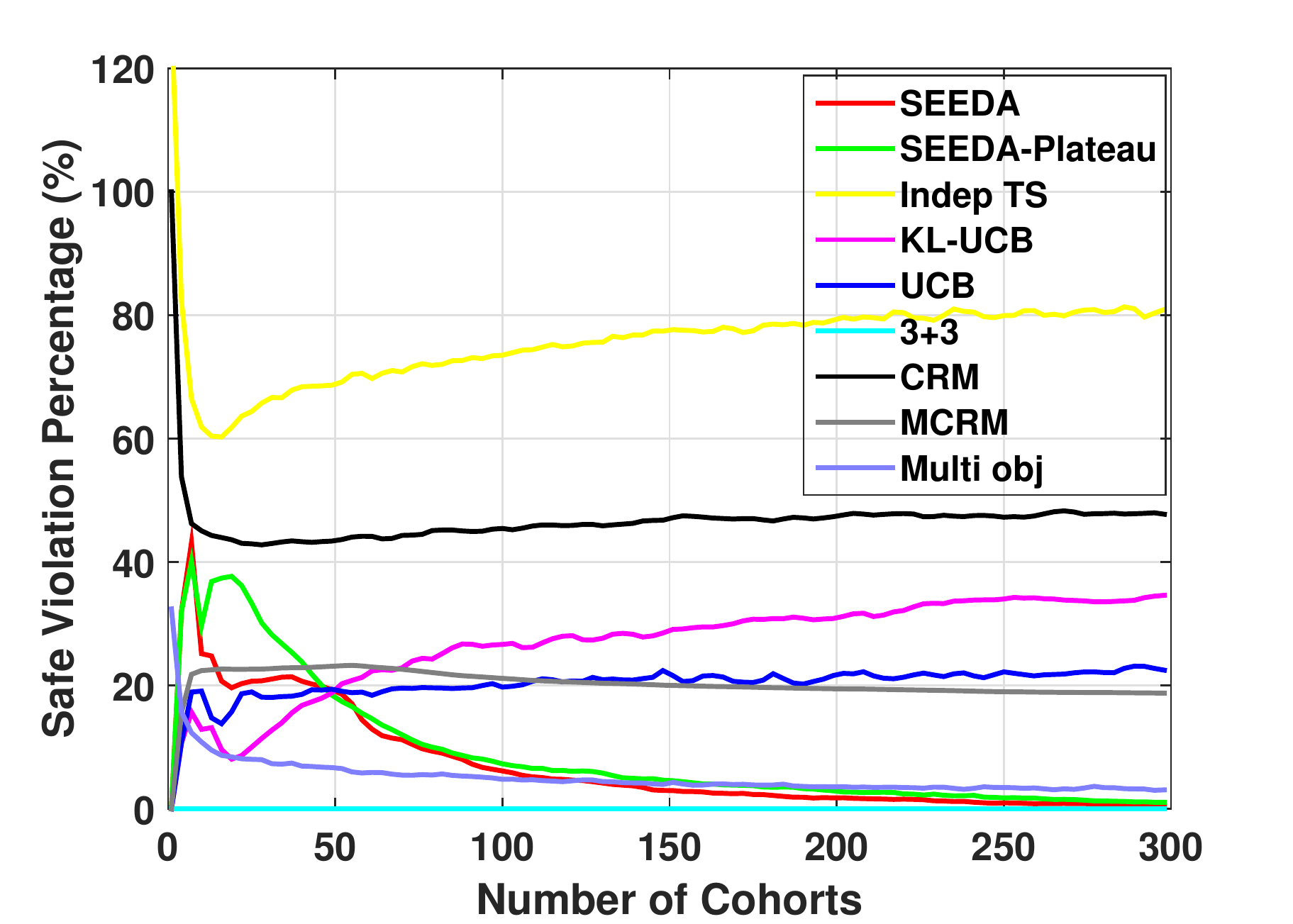}}  }
        }
    \caption{Comparison of efficacy per patient (left) and the safety violation percentage (right).}
    \label{fig:effvio2}
\end{figure}

To have a deeper understanding of the tradeoff between efficacy and toxicity, we plot side-by-side the convergence of efficacy and toxicity as $t$ increases in Fig.~\ref{fig:effvio2}. KL-UCB, UCB and Independent TS have good convergence but suffer from significant safety violation in the process since they do not consider the safety constraint during exploration. CRM has higher efficacy at the cost of bad safety constraint violation, while 3+3 performs poorly in efficacy but has the lowest safety probability; this behavior is similarly observed for multi-objective bandits. The SEEDA(-Plateau) algorithm, in comparison, converges to the optimal efficacy at a slower rate, but the exploration process is carefully controlled so that the safety violation is minimized, which is evident from the right subplot of Fig.~\ref{fig:effvio2}.

\subsubsection{Sample efficiency}
\label{sec:sim_eff}

\begin{figure}
\centering
\centerline{\includegraphics[width=0.4\textwidth]{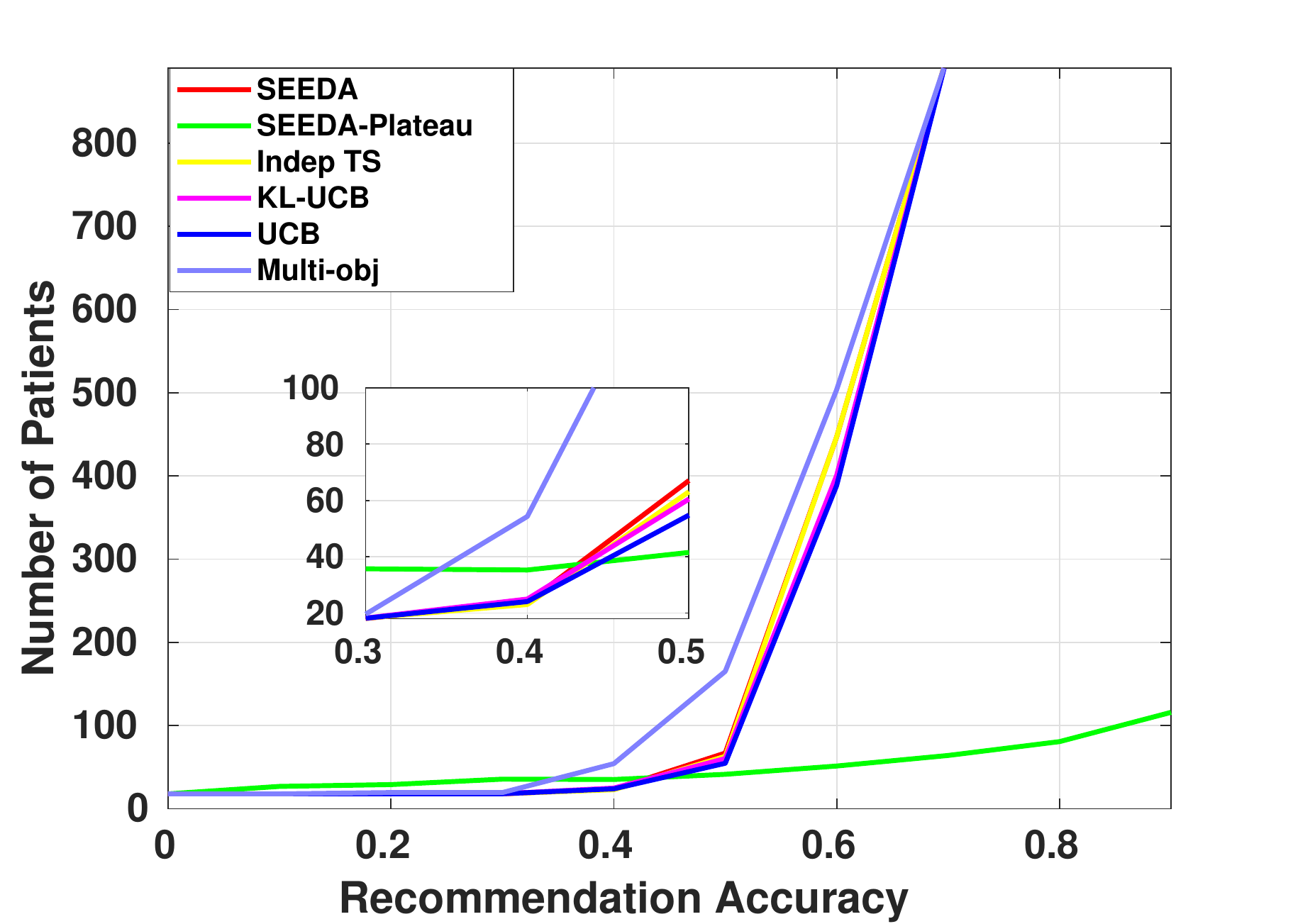}}
\caption{The minimum number of trial participants to achieve a given a recommendation accuracy.}
\label{fig:earlystop}
\end{figure}

Sample efficiency is measured by the minimum number of trial participants to achieve a pre-specified recommendation accuracy (also known as \emph{early stopping}  \cite{Montori2005}). We start the trial with a minimum of 6 patients, and continue recruiting patients until the stopping condition is triggered.  Fig.~\ref{fig:earlystop} plots the average minimum number of patients to achieve a given a recommendation accuracy for different algorithms\footnote{3+3, CRM and MCRM are excluded since they only target finding MTD.}.  We see that SEEDA-Plateau outperforms all other algorithms by a large margin, thanks to the ``double dipping'' of the model assumptions which gives the most accurate estimation of the optimal dose. In comparison, SEEDA performs similarly to the baseline algorithms. The reason is that the goal of SEEDA is to recommend the efficacy-maximal dose that satisfies the safety constraint. In this particular setting, both dose 3 and 4 satisfy this condition, and SEEDA does not have the mechanism to further minimize toxicity. This leads to a recommendation error that is similar to other baseline designs.

The  sample efficiency advantage of SEEDA-Plateau is of critical importance in practice, as the significant cost associated with clinical trials is mostly proportional to the number of trial participants. Furthermore, reducing the number of patients while achieving the same level of accuracy minimizes the safety and ethical concern in the trial, which is another important consideration.

\subsection{Real-world datasets}

\begin{table*}[h]
\caption{Recommendation \& allocation percentages of the neurodeg dataset. In each cell the first row reports the mean value over 1000 repetitions, and the second row reports the (standard deviation).}
\label{tab:IBS}
\vspace{-0.1in}
\begin{center}
\adjustbox{max width=0.99\textwidth}{
\begin{tabular}{ l || c | c | c |c | c  |||  c | c | c |c | c  }
\hline
{} & \multicolumn{5}{c}{Recommended} & \multicolumn{5}{c}{Allocated} \\
\Xhline{3\arrayrulewidth}
Toxicity & 0.01 &0.08& \textbf{0.30}& 0.60 &0.80 &
  0.01 &0.08& \textbf{0.30}& 0.60 &0.80\\
\hline
Efficacy &0.01 &0.35 & \textbf{0.45}& 0.52 &0.57&
0.01 &0.35 & \textbf{0.45}& 0.52 &0.57\\ 
\Xhline{3\arrayrulewidth} 
\multirow{2}{*}{SEEDA} & 0.60& 32.91& 66.14& 0& 0&                         5.58& 34.14& 59.60& 0.33& 0.33 \\
&(0.40) &(10.57)&(10.59) &(0)&(0) &(0.42) &(6.08)&(6.25) &(0.25) &(0.01)
\\
\hline
\multirow{2}{*}{SEEDA-Plateau} & 0.99& 32.66& 66.00& 0& 0 &               5.09& 35.02& 59.21& 0.33& 0.33  \\
&(0.31) &(10.12) &(10.36) &(0) &(0) &(2.05) &(7.78) &(6.64) &(0.02) &(0)
\\
\hline
\multirow{2}{*}{Independent TS} & 3.39& 51.28& 44.47& 0.38& 0.46&               0.80& 3.50& 7.59& 23.37& 64.68  \\
&(2.56) &(9.92) &(10.34) &(0.33) &(0.37) &(0.60) &(2.70) &(5.21) &(10.19) &(12.51)
\\
\hline
\multirow{2}{*}{KL-UCB}  &0.07& 55.74& 28.67& 5.43& 0.07&                   98.0& 0.42& 0.47& 0.52& 0.55 \\
&(0.06) &(12.38) &(12.76) &(2.10) &(0.05) &(2.62) &(0.23) &(0.28) &(0.49) &(0.04)
\\
\hline
\multirow{2}{*}{UCB} &0.81& 41.68& 57.24& 0.23& 0.01&                6.88& 15.09& 20.10& 25.75& 32.16
\\&(0.74)&(16.07)&(16.07)&(0.17)&(0.01)&(0.29)&(1.60)&(2.10)&(2.55)&(2.87)
\\
\hline
\multirow{2}{*}{3+3} & 0& 2.40& 12.00& 17.60& 45.20&                               16.04& 17.82& 20.19& 18.12& 16.81 \\&(0)&(1.02)&(2.35)&(3.44)&(10.34)&(5.60)&(9.48)&(1.84)&(1.60)&(4.20)
\\
\hline
\multirow{2}{*}{CRM} & 0& 0& 0& 100& 0&  0& 0& 0& 99.66& 0.33 \\&(0)&(0)&(0)&(0)&(0)&(0)&(0)&(0)&(0.01)&(0.01)
\\
\hline
\multirow{2}{*}{MCRM} &4.33& 26.47& 69.18& 0& 0&                       4.67& 26.40& 68.92& 0& 0\\&(0.25)&(1.80)&(1.86)&(0)&(0)&(0.25)&(1.80)&(2.10)&(0)&(0)
\\
\hline
\multirow{2}{*}{Multi-obj}& 0.24& 15.33& 17.59& 0.12& 0.03&                 24.33& 26.12& 18.95& 16.05& 14.52 \\&(0.13)&(9.65)&(9.71)&(0.05)&(0.03)&(4.28)&(3.51)&(6.11)&(2.93)&(2.61) 
\\
\hline
\end{tabular}
}
\end{center}
\vspace{-0.1in}
\end{table*}

\begin{table*}[h]
\caption{Recommendation \& allocation percentages of the IBScovars datasets. In each cell the first row reports the mean value over 1000 repetitions, and the second row reports the (standard deviation).}
\label{tab:neu}
\vspace{-0.1in}
\begin{center}
\adjustbox{max width=0.99\textwidth}{
\begin{tabular}{ l || c | c | c |c | c  |||  c | c | c |c | c  }
\hline
{} & \multicolumn{5}{c}{Recommended} & \multicolumn{5}{c}{Allocated} \\
\Xhline{3\arrayrulewidth}
Toxicity probabilities & 0.01 &0.10& \textbf{0.30}& 0.70 &0.95 &
  0.01 &0.10& \textbf{0.30}& 0.70 &0.95\\
\hline
Efficacy probabilities &0.01 &0.20& \textbf{0.27}& 0.33 &0.43&
0.01 &0.20& \textbf{0.27}& 0.33 &0.43\\ 
\Xhline{3\arrayrulewidth} 
\multirow{2}{*}{SEEDA} &  1.14 &	35.04&	63.47& 0&	0&
 10.11 &	34.35 &	54.86 &0.33&	0.33\\&(1.31)&(7.58)&(7.60)&(0)&(0)&(0.82)&(5.42)&(5.57)&(0.17)&(0.01) 
 \\
\hline
\multirow{2}{*}{SEEDA-Plateau} &2.08&	36.51&	61.06&	0 &	0
& 8.97 &	34.60&	55.75&	0.33&	0.33\\&(2.76)&(10.31)&(10.42)&(0)&(0)&(4.00)&(4.34)&(7.83)&(0.02)&(0.01)
\\
\hline
\multirow{2}{*}{Independent TS}&7.52& 48.47&	43.30&	0.31&	0.39 &
 1.82 &	3.89&	26.66& 	20.65 &	23.17 \\&(7.15)&(9.79)&(9.71)&(0.60)&(0.17)&(0.85)&(3.61)&(10.74)&(13.54)&(10.05)
 \\
\hline
\multirow{2}{*}{KL-UCB}  &28.55&	44.90&	23.24&	3.28&	0&
98.33&	0.37&	0.40&	0.42&	0.45 \\
&(9.21)&(9.95)&(10.06)&(2.60)&(0) &(0.35)&(0.44)&(0.87)&(0.06)&(0.60)
\\
\hline
\multirow{2}{*}{UCB}&1.73 &	45.26& 52.81& 0.17&	0.01&
9.41 &	15.37&	18.94&	23.22&	33.04\\
&(1.21)&(9.21)&(9.25)&(0.08)&(0.01)&(0.38)
&(1.45)&(1.87)&(2.26) &(2.61)\\
\hline
\multirow{2}{*}{3+3} &  	2.40& 	12.00&	17.60& 45.20& 22.80&
16.04 &	17.82 &  20.19&	18.12 &	22.81\\&(0.88)&(7.65)&(6.87)&(6.86)  &(8.87)&(2.85)&(5.29)&(8.29)&(5.52)&(5.45)
\\
\hline
\multirow{2}{*}{CRM} & 0	&0&	0&	1.35&	98.65 &	  
0	&0&0 &	99.66 &	0.33\\&(0)&(0)&(0 )&(0.10)&(0.04)&(0)&(0)&(0)&(0.86) &(0.03)
\\
\hline
\multirow{2}{*}{MCRM} & 4.34&	26.91&	68.74&	0&	0
&4.67 &	26.83 &	68.49 &0&	0\\&(0.26)&(2.15)&(2.20)&(0)&(0)&(0.04)&(0.05)&(0.91)&(0)&(0)
\\
\hline
\multirow{2}{*}{Multi-obj}& 0.45&	16.18&	16.56&	0.10 &	0.02& 
 1.23 &    3.27&    5.79&   13.11 &   76.56 \\&(0.25)&(5.49)&(10.53)&(0.03)&(0) &(1.14)&(3.12)&(5.12)&(6.43)&(6.04)
 \\
\hline

\end{tabular}
}
\end{center}
\vspace{-0.1in}
\end{table*}

We evaluate the SEEDA algorithms in two real-world datasets \emph{neurodeg} and \emph{IBSCovars} based on \cite{Biesheuvel2002}.  We first extract \ssf{dose} and \ssf{resp} variables from the observations reported in the dataset. With these samples, we fit them into a commonly used Emax dose-response model as in \cite{Bornkamp2011} with an R package implementation provided by \cite{Yoshida2019}. The resulting models are as follows.
\begin{align*}
    \text{neurodeg:}& \quad \ssf{resp}=169.94+\frac{12.95 \ssf{dose}}{1.85+\ssf{dose}}, \\
    \text{IBScovars:}& \quad \ssf{resp}=0.26+\frac{0.68 \ssf{dose}}{4.01+\ssf{dose}}.\end{align*}
As for the toxicity event, since it is not reported in the dataset, we resort to simulations with model \eqref{eqn:toxicity}. 

The allocation and recommendation percentages of each dose for all the algorithms are shown in Table \ref{tab:IBS} and Table \ref{tab:neu} for both datasets. We have similar observations as in the synthetic experiment that SEEDA and SEEDA-Plateau recommend the correct doses majority of the times, while the suboptimal recommendation is mostly safe in that the doses immediately below MTD are recommended second most. The same is true for allocation.

\section{Related works}
\label{sec:related}

This work is concerned with adaptive phase I clinical trials, whose uptake in practice is starting to increase considerably.  See \cite{Bretz2017, Pallmann2018} for recent comprehensive surveys. The main motivation to use these adaptive designs is to learn as the trial progresses and use this learning to deliver more efficient or more ethically appealing trials. Adaptive clinical trial with sequential patient recruitment is considered in \cite{Atan2019a}, but it does not address the subsequent dose allocation.  The 3+3 and the CRM designs or their variations remain the de facto adaptive designs in practice for dose-finding studies \cite{Petroni2017,Pallmann2018}, although new methodologies that aim at better safety protection are also proposed \cite{Lee2019}. In recent years, there is a growing interest in adaptive trial designs for MTA because of its different dose-response relationships \cite{Zang2014,Riviere2017}, but these studies do not explicitly enforce the safety constraints during the trial; neither do they provide theoretical guarantees on the trial performance.

Multi-armed bandit has long been considered as an important tool for learning in clinical trials, dating back to the earliest papers of \cite{Thompson1933,Robbins1952}. Developing bandit models and algorithms that better suit the specific requirements of adaptive clinical trials has attracted some attention in recent years. Villar et. al \cite{Villar2015,Villar2018a} adopted the (modified) forward-looking Gittins index rule for multi-arm clinical trials. The authors of \cite{Wang2018aistats} propose a regional bandit model that can be applied to learning the drug dosage and patient response relationship.  The sample complexity of thresholding bandit is analyzed in \cite{Garivier2017}, which matches MTD identification. Furthermore, dose-finding clinical trials with heterogeneous groups are investigated in \cite{lee2020contextual} from a MAB perspective. Probably the closest work to ours is \cite{Aziz2019}, which also considers both toxicity and efficacy. However, the safety constraint, which is an essential constraint of real-world phase I trials, has not been explicitly considered in these papers.

On the other hand, the problem of safe exploration has attracted a lot of attention recently, albeit often in control \cite{Koller2018} and general reinforcement learning \cite{Berkenkamp2017}. The authors in \cite{Sui2015} propose the SAFEOPT algorithm for safe exploration in Gaussian processes, and \cite{Kazerouni2017} presents a variant of linear UCB method for the contextual linear bandit problem. A different line of works \cite{Maillard2013,Galichet2013} consider minimizing risk in MAB, but they are mostly casted in the mean-variance framework with respect to the reward distribution.

\section{Conclusions}
\label{sec:conc}

Learning in adaptive clinical trials faces several unique challenges that have not been well addressed, which may have contributed to their lack of adoption in actual clinical trials.  In particular, the safety constraints resulting from ethical and societal considerations have been insufficiently researched, which has motivated us to develop the SEEDA algorithm that explicitly imposes safety constraints (in terms of toxicity) while also aiming for maximum patient response in a dose-finding study. Theoretical analysis of SEEDA is carried out and the proposed algorithm is further extended to the increase-then-plateau efficacy model and shown to have smaller regret thanks to the unimodal structure. The performance advantages over state-of-the-art adaptive clinical trial designs are illustrated with experiments on both synthetic and real-world datasets. 

\section{Acknowledgements}
CS acknowledges the funding support from Kneron, Inc. SSV thanks the funding received from the National Institute for Health Research Cambridge Biomedical Research Centre at the Cambridge University Hospitals NHS Foundation Trust and the UK Medical Research Council (grant number: MC\_UU\_00002/3). The research of MV has been supported by ONR and NSF 1524417 and 1722516.
\medskip


\bibliographystyle{icml2020}
\bibliography{dose}

\newpage
\onecolumn
\appendix

\vbox{
\begin{center}
{\noindent {\Large{\bf Supplementary Material: Learning for Dose Allocation in Adaptive Clinical Trials with Safety
Constraints}}}
\end{center}
}

\begin{center}
\large{{Cong Shen}, Zhiyang Wang, Sof{\'i}a S.~Villar,  Mihaela van der Schaar}
\end{center}

\vspace{0.1in}

\section{Preliminaries}
\label{appx:assump}

Before presenting the technical proofs, we introduce some notations and regularity assumptions on the dose-toxicity model, which can be verified to hold for Eqn.~\eqref{eqn:toxicity}.  For a general toxicity function $p_k(a)$ of an unknown parameter $a\in\mathcal{A}$, the following regularities are imposed:
\begin{assumption}
\label{ass:1}
\begin{enumerate}
\item[1)] \textbf{Monotonicity:} For each $k \in \mathcal{K}$ and $a,a' \in \mathcal{A}$ there exists
$C_{1,k} > 0$ and $1 < \gamma_{1,k}$, such that
$
|p_k(a)-p_k(a')| \geq C_{1,k}|a-a'|^{\gamma_{1,k}}
$.
\item[2)] \textbf{H\"{o}lder continuity:} For each $k \in \mathcal{K}$ and $a,a' \in \mathcal{A}$ there exists
$C_{2,k} > 0$ and $0 < \gamma_{2,k}\leq 1$, such that
$
|p_k(a)-p_k(a')| \leq  C_{2,k}|a-a'|^{\gamma_{2,k}}
$.
\end{enumerate}
\end{assumption}
We note that both monotonicity and continuity assumptions are mild and standard in the literature; see \cite{Wang2018aistats}. 
Proposition~\ref{prop:1} immediately follows with Assumption \ref{ass:1}.
\begin{proposition}
\label{prop:1}
For functions $p_k(a), \forall k \in \mathcal{K}$ that satisfy Assumption \ref{ass:1}, we have:
\vspace{-0.1in}
\begin{enumerate}
\item[1)] $p_k(a)$ is invertible;
\item[2)] For each $k \in \mathcal{K}$ and $d, d' \in \mathcal{P}$, we have
$|p_{k}^{-1}(d)-p_{k}^{-1}(d')| \leq \bar{C}_{1,k}|d-d'|^{\bar{\gamma}_{1,k}}$,
where $\bar{\gamma}_{1,k} = \frac{1}{\gamma_{1,k}}$, $\bar{C}_{1,k} = {(\frac{1}{C_{1,k}})}^{\frac{1}{\gamma_{1,k}}}$.
\end{enumerate}
\vspace{-0.1in}
\end{proposition}
For ease of exposition, we denote $C_1=\min C_{1,k}$, $C_2=\max C_{2,k}$, $\gamma_1=\max \gamma_{1,k}$, $\gamma_2=\min \gamma_{2,k}$, $\bar{\gamma}_1=1/\gamma_1$, and $\bar{C}_1=C_1^{-\bar{\gamma}_1}$.

\section{Select Design Parameters}
\label{appx:chooseparam}

The parameters appeared in Assumption {\ref{ass:1}} collectively determine the confidence interval in Eqn.~\eqref{eqn:alpha}. We take function \eqref{eqn:toxicity} as an example to show how to select these parameters. We have
\begin{align*}
    |p_k(a)-p_k(a')| &\geq C_{1,k}|a-a'|^{\gamma_{1,k}},\\
    \frac{|p_k(a)-p_k(a')|}{|a-a'|} &\geq C_{1,k}|a-a'|^{\gamma_{1,k}-1},\\
    \min_{a\in\mathcal{A}}{p'_k(a)} &\geq C_{1,k} |\mathcal{A}|^{\gamma_{1,k}-1},\\
    \log\left(\frac{\tanh(d_k)+1}{2}\right) &\geq C_{1,k} |\mathcal{A}|^{\gamma_{1,k}-1}.
\end{align*}
Therefore, we can first set $\gamma_{1,k}$ as $\frac{3}{2}$ and find the corresponding $C_{1,k}$. Then, with the known function $p_k(a)$, parameters can be approximately calculated.

\section{Proof of Lemma~\ref{lem:1}}
\label{appx:lem:1}
\begin{align}
     \nonumber\mathbb{P}[\hat{a}(t)+\alpha(t) < p^{-1}_i(\theta)]&\leq \mathbb{P}[\hat{a}(t) +\alpha(t) < a^*] \\
     & \leq\mathbb{P}[|a^* - \hat{a}(t)| > \alpha(t)] \nonumber \\
     \nonumber &\leq \mathbb{P}\left[\sum\limits_{k=1}^K w_k(t-1)\bar{C}_1|\hat{p}_k(t)-p_k(a^*)|^{\bar{\gamma}_1}>\alpha(t)\right]\\
    \nonumber&\leq \sum\limits_{k=1}^K \mathbb{P}\left[|\hat{p}_k(t)-p_k(a^*)|>\left(\frac{\alpha(t)}{w_k(t-1)\bar{C}_1K}\right)^{\gamma_1}\right]\\
    \label{eqn:lemma1-1}&\leq \sum\limits_{k=1}^K 2\exp\left(-2N_k(t) \left(\frac{\alpha(t)}{w_k(t)\bar{C}_1K}\right)^{2\gamma_1}\right)\\
    \label{eqn:lemma1-2}& \leq 2K\exp\left(-2 \left(\frac{\alpha(t)}{\bar{C}_1 K}\right)^{2\gamma_1}t\right)=\delta.
\end{align}
Inequality \eqref{eqn:lemma1-1} is from the Hoeffding's inequality and \eqref{eqn:lemma1-2} is derived from the definition of $N_k(t)=t w_k(t)$ and Assumption \ref{ass:1} with  $\gamma_1>1$.

\section{Proof of Lemma~\ref{lem:2}}
\label{appx:lem:2}
From the Hoeffding's Inequality and  Eqn.~\eqref{eqn:ti}, we have:
\begin{align*}
    \alpha(t)\leq p_k^{-1}(\theta)-a^*-\epsilon=\Delta_k-\epsilon,
\end{align*}
where $\Delta_k=|a^*-p_k^{-1}(\theta)|$ denotes the gap between the true value of parameter $a$ and the parameter corresponding to when the toxicity of dose level $d_k$ is exactly at the MTD threshold $\theta$. When $t>t_1$ and with the definition of $\alpha(t)$ in Eqn.~\eqref{eqn:alpha}, the lemma can be immediately derived.

\section{Proof of Theorem~\ref{thm:1}}
\label{appx:thm:1}
Depending on whether the optimal dose level is included in the admissible set or not, we can decompose the regret into two parts:
\begin{align*}
    R(n)&=\sum\limits_{t=1}^n\mathbb{P}[k^*\notin \mathcal{D}_1(t)]Q +\mathbb{P}[k^*\in \mathcal{D}_1(t)]R_2(n)\\
    & \leq {n}\delta Q +R_2(n).
\end{align*}
The probability of the first error event $\{k^*\notin \mathcal{D}_1(t)\}$ can be bounded by Lemma \ref{lem:1}, which indicates that at each step $t$ the probability of a safe dose level being excluded from the admissible set is bounded by $\delta$. For the second part, $R_2(n)$ represents the regret when the optimal dose  is included in the admissible set. In this case, the error event is due to the inaccuracy of parameter estimation at the beginning as well as the limited efficacy information provided by each sample. Using Lemma \ref{lem:2}, we have:
\begin{align*}
    R_2(n)&\leq t_1+(K-M)\sum\limits_{t=1}^n \exp(-2t\epsilon^2)+\sum\limits_{t=t_1+1}^n\sum\limits_{d_k:p_k\leq \theta}\mathbbm{1}\{I(t)=k\}\\
&\leq t_1+\frac{K-M}{2\epsilon^2}+ \sum\limits_{d_k:p_k\leq \theta}\frac{ c \log(n)}{q^*-q_k}.
\end{align*}
Putting the regret from both error events together leads to \eqref{eqn:regbnd}, which completes the proof.

\section{Proof of Theorem~\ref{thm:2}}
\label{appx:thm:2}
First we note:
\begin{eqnarray*}
p_{I(t)}(a^*)-\theta  &\leq& p_{I(t)}(a^*)-\theta+\theta-p_{I(t)}(a^*-\alpha(t)) \\
   & \leq & C_2|a^*-\hat{a}(t)+\alpha(t)|^{\gamma_2}.
\end{eqnarray*}
Thus, the probability can be upper bounded as:
\begin{equation*}
    \mathbb{P}[\hat{a}(t)-a^*>\alpha(t)+\epsilon] \leq \exp(-2t(\alpha(t)+\epsilon)^2).
\end{equation*}
Reorganizing the terms, we finally have
\begin{equation*}
    \mathbb{P}\left[\frac{1}{n} \sum\limits_{t=1}^n p_{I(t)}(a^*)-\theta<C_2\epsilon^{\gamma_2}\right] \geq 1-\exp(-2t(\alpha(t)+\epsilon)^2) \geq 1-\delta.
\end{equation*}

\section{Proof of Corollary~\ref{cor:accuracy}}
\label{appx:cor:accuracy}

\begin{align*}
    \mathbb{P}[|\hat{a}(n)-a^*|\geq \Delta_{M}]& \leq \sum\limits_{k=1}^K \mathbb{P}\left[|\hat{p}_k(t)-p_k(a^*)|>\left(\frac{\Delta_{M}}{w_k(t)\bar{C}_1K}\right)^{\gamma_1}\right]\\
    &\leq \sum\limits_{k=1}^K 2\exp\left(-2N_k(n) \left(\frac{\Delta_{M}}{w_k(t)\bar{C}_1K}\right)^{2\gamma_1}\right)\\
    & \leq 2K\exp\left(-2 \left(\frac{\Delta_{M}}{\bar{C}_1 K}\right)^{2\gamma_1}n\right).
\end{align*}

\section{Proof of Theorem~\ref{thm:3}}
\label{appx:thm:3}
We first establish Lemma~\ref{lem:3}, whose proof directly follow Theorem C.1 in \cite{Combes2014}.
\begin{lemma}
\label{lem:3}
$\mathbb{E}[l_k(n)]= O(\log(\log(n)))$, for each $k\neq k^*$.
\end{lemma}
Then, following the similar proof steps in Theorem~\ref{thm:1}, we have the bound in \eqref{eqn:thm3}.

\section{Proof of Theorem~\ref{thm:4}}
\label{appx:thm:4}

Since $k^*=\min\{M,N\}$ and $L_1(n)$ and $L_2(n)$ are the estimations for $N$ and $M$ respectively, $\{\hat{d}_r(n)\neq k^*\}\subseteq E_1\bigcup E_2$, where $E_1=\{L_1(n)\neq N\}$, $E_2=\{L_2(n) \neq M\}$. The latter can be bounded by Corollary \ref{cor:accuracy}. With the notation $\beta_k(n)=\sqrt{\frac{c\log(n)}{N_k(n)}}$, the probability of $E_1$ can be bounded as follows:
\begin{align*}
\mathbb{P}[L_1(n)< M]&\leq \mathbb{P}[ |\hat{q}_{N}(n)-\hat{q}_{N-1}(n)|\leq \beta_{N-1}(n)+\beta_{N}(n) ]\\
&\leq \mathbb{P}[\hat{q}_{N-1}(n)-q_{k}+q_{N}-\hat{q}_{N}(n)\leq q_N-q_{N-1}-\beta_{N-1}(n)-\beta_{N}(n)]
\\ &\leq 2\exp\left(-2N_{N-1}(n)\left(\frac{q_N-q_{N-1}-\beta_{N-1}(n)-\beta_{N}(n)}{2}\right)^2\right)
\\ &\leq 2\exp\left(-2f(N-1) \log(n) \left(\frac{\Delta_{N-1,N}-\beta_{N-1}(n)-\beta_{N}(n)}{2}\right)^2\right)\\
&=o\left(n^{-\frac{5}{2}}\right).
\end{align*}
Furthermore, 
\begin{align*}
\mathbb{P}[L_1(n)> M]&\leq \mathbb{P}[|\hat{q}_N(n)-\hat{q}_{N+1}(n)|>\beta_{N}(n)+\beta_{N+1}(n)]\\
& \leq\mathbb{P}[|\hat{q}_N(n)-q_N|+|q_{N+1}-\hat{q}_{N+1}(n)|> \beta_{N}(n)+\beta_{N+1}(n) ]\\
& \leq \mathbb{P}[|\hat{q}_N(n)-q_N|> \beta_{N}(n)]+\mathbb{P}[|\hat{q}_{N+1}(n)-q_{N+1}|> \beta_{N+1}(n)]\\
&\leq \frac{2}{n^c}.
\end{align*}
Lastly, $f(N-1)$ is the coefficient of the lower bound of $N_{N-1}(n)$, and can be written as (see Theorem 4.1 in \cite{Combes2014})
\begin{align*}
    f(N-1)=\frac{1}{I(q_{N-1},q_N)}.
\end{align*}
This completes the proof.

\section{Baseline designs in the experiments}
\label{appx:sim:baseline}

The following baseline designs are used for comparison to SEEDA and SEEDA-Plateau in the experiments.


\begin{itemize}
\item \textbf{KL-UCB} \cite{Garivier2011}: This approach ignores the safety constraint and focuses entirely on efficacy during allocation, as for each patient it allocates the dose level with the highest efficacy index. The efficacy performance for each dose level is characterized by the KL-UCB index. 
However, at the end of the experiment, a dose level is recommended according to $\hat{d}(n)=\arg\max_{k:\hat{p}_k(n)\leq \theta}\hat{q}_k(n)$, where $\hat{q}_k(n)$ and $\hat{p}_k(n)$ are the last empirical estimations of toxicity and efficacy for dose level $d_k$. This suggests that safety constraint is considered in recommendation.
Accordingly, type I and type II errors are defined as:
    \begin{gather*}
    e_1=\sum_{k\in\mathcal{K}}\mathbbm{1}\{p_k\leq \theta\}\mathbbm{1}\{\hat{p}_k(n)>\theta\},\\
    e_2=\sum_{k\in\mathcal{K}}\mathbbm{1}\{p_k> \theta\}\mathbbm{1}\{\hat{p}_k(n)\leq \theta\}.
    \end{gather*}

\item \textbf{UCB-1} \cite{Auer2002}: The allocation and recommendation rules are similar to KL-UCB above, with the only difference that the dose level with the highest UCB-1 index of efficacy is allocated to the patient.

\item \textbf{Independent Thompson Sampling (TS)} \cite{Thompson1933, Aziz2019}: Toxicity and efficacy are estimated with Bayesian indices: $$\tilde{p}_k(t)\sim Beta(S^p_k(t)+1,N_k(t)-S^p_k(t)+1),$$ and $$\tilde{q}_k(t)\sim Beta(S^q_k(t)+1,N_k(t)-S^q_k(t)+1),$$ where $S^p_k(t)$ counts the number of toxic outcomes of dose level $k$ among the first $t$ patients and $S^q_k(t)$ counts the number of effective responses. The dose with maximum $\tilde{q}_k(t)$ is allocated to the $t$-th patient and $\hat{d}(n)=\arg\max_{k:\tilde{p}_k(n)\leq \theta}\tilde{q}_k(n)$ is recommended. Definitions of type I and type II errors are slightly modified to:
    \begin{gather*}
    e_1=\sum_{k\in\mathcal{K}}\mathbbm{1}\{p_k\leq \theta\}\mathbbm{1}\{\tilde{p}_k(n)>\theta\},\\
    e_2=\sum_{k\in\mathcal{K}}\mathbbm{1}\{p_k> \theta\}\mathbbm{1}\{\tilde{p}_k(n)\leq \theta\}.
    \end{gather*}

\item \textbf{CRM} \cite{OQuigley1990}: We here employ the CRM algorithm with the same one-parameter toxicity model in our paper:
\begin{equation*}
p_k(a)=\left(\frac{\tanh(d_k)+1}{2}\right)^a.
\end{equation*}
We choose a typical prior distribution as $a\sim \exp(0.5)$. Therefore, $d_k$ can be solved with $prior_{tox}$ and the prior mean of $a$. $\pi_t(a)$ denotes the posterior distribution of $a$ after observing the outcomes of the first $t$ patients. The allocation rule is a greedy one:
\begin{eqnarray*}
I^{CRM}_t &=& \arg\min\limits_{k\in\mathcal{K}} |\theta-p_k(\hat{a}(t))|, \\
\hat{a}(t)&=&\int_0^\infty {a \text{d} \pi_t(a)},
\end{eqnarray*}
where $\hat{a}(t)$ is the posterior mean value. With this estimation, the final recommendation rule can be written as: $$\hat{d}(n)=\arg\min_{k\in\mathcal{K}}|\theta-p_k(\hat{a}(n))|.$$

\item \textbf{3+3} \cite{Storer1989}: The lowest dose is first given to 3 patients. If none reports a toxic outcome, the next lowest dose level is given to the next 3 patients. If there are less than 2 among these 6 patients who report toxic outcome, the next lowest dose level is given to the next 3 patients; otherwise the experiment is stopped and the dose level used before stopping is recommended as MTD.

\item \textbf{MCRM} \cite{Neuen2008}: This algorithm classifies the probability of toxicity into four categories. For our simulated setting, the categories are set as:
\begin{align*}
    &\text{Under-dosing:} & & \pi_a(d)\in(0,0.20]\\
    &\text{Targeted toxicity:} & & \pi_a(d)\in(0.20,0.35]\\
    &\text{Excessive toxicity:} & & \pi_a(d)\in(0.35,0.60]\\
    &\text{Unacceptable toxicity:} && \pi_a(d)\in(0.60,1.00]
\end{align*}
The recommendation and the allocation rules are to maximize the probability of targeted toxicity while controlling the probability of excessive or unacceptable toxicity at $P^{thre}=25 \%$. Based on the posterior distribution of the toxicity, the probability that the toxicity falls in the above four categories can be calculated. The probability that it falls in Targeted category is denoted as $P_i^t$ while falls in Excessive and Unacceptable categories as $P_i^e$. The selection rule is therefore $I_t = \arg\max\limits_{P_i^e\leq P^{thre} } P_i^t$.

\item \textbf{Multi-objective Bandits} \cite{Yahyaa2015}: We implement the Pareto Thompson Sampling algorithm of \cite{Yahyaa2015} in our experiments. Specifically, after getting the estimations of toxicity and efficacy of each dose from running the Independent TS design, the algorithm computes the Pareto optimal dose level set $\mathcal{I}^*$, which means $\forall i \in \mathcal{I}^*,\forall j \notin \mathcal{I}^*, \tilde{p}_i(t)\leq \tilde{p}_j(t)  $ or $\tilde{q}_i(t)\geq \tilde{q}_j(t)$.
\end{itemize}

Other policies designed for MTA, such as MTA-RA, depend on a different truncated two-parameter logistic efficacy model \cite{Riviere2017}. In our setting, the exact efficacy model is assumed to be unknown -- we only make the increase-then-plateau assumption.


\section{Additional experiment results under the same setting as in Section~\ref{sec:sim}}
\label{appx:sim:setting2}

Due to space limitations, we were not able to include all the experiment results of the setting in Section~\ref{sec:sim}. These additional results are provided here.

\begin{figure}
\centerline{
        \subfigure{
        { \includegraphics[width=0.48\textwidth]{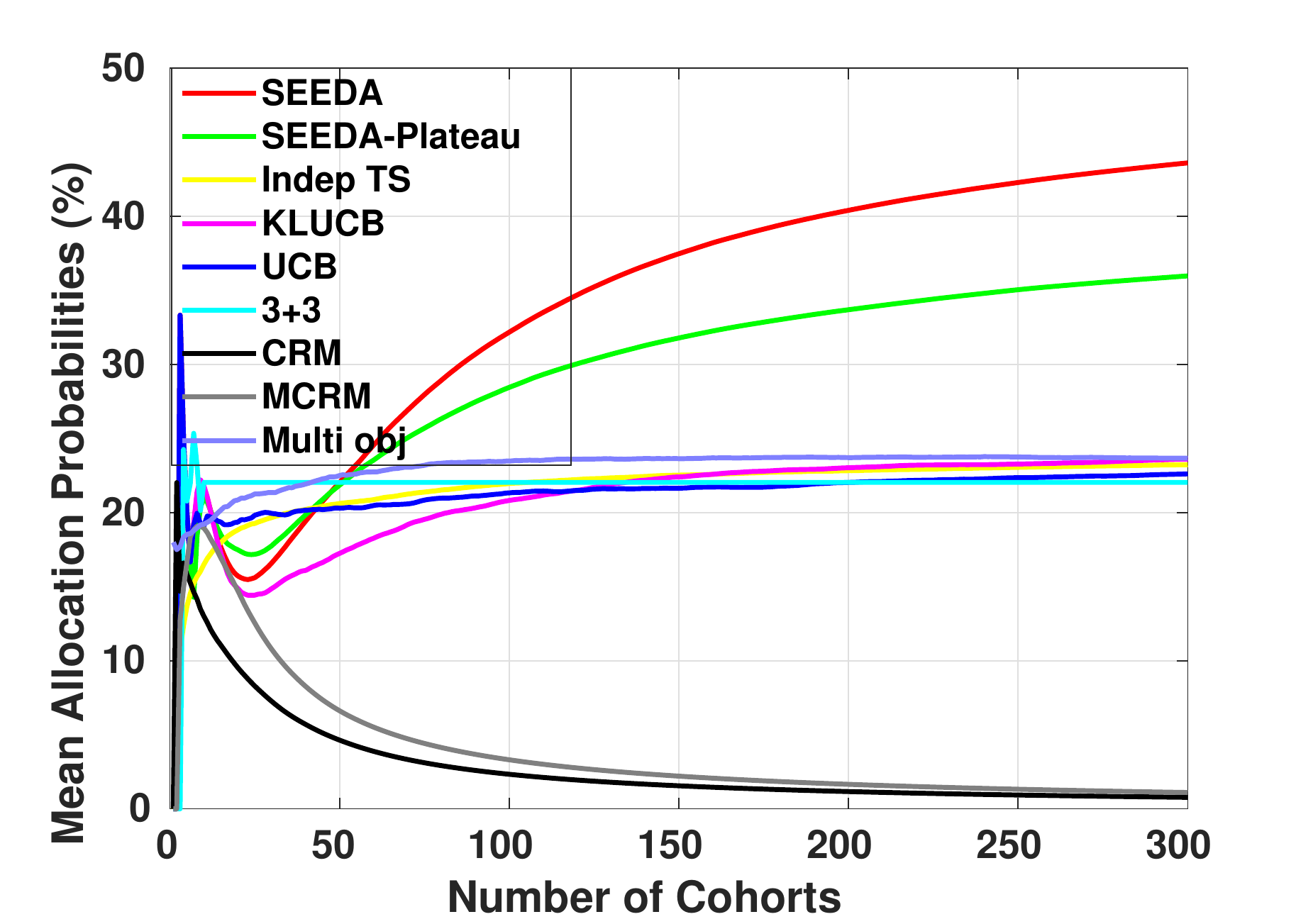}} }
        \hfil
        \subfigure{
        { \includegraphics[width=0.48\textwidth]{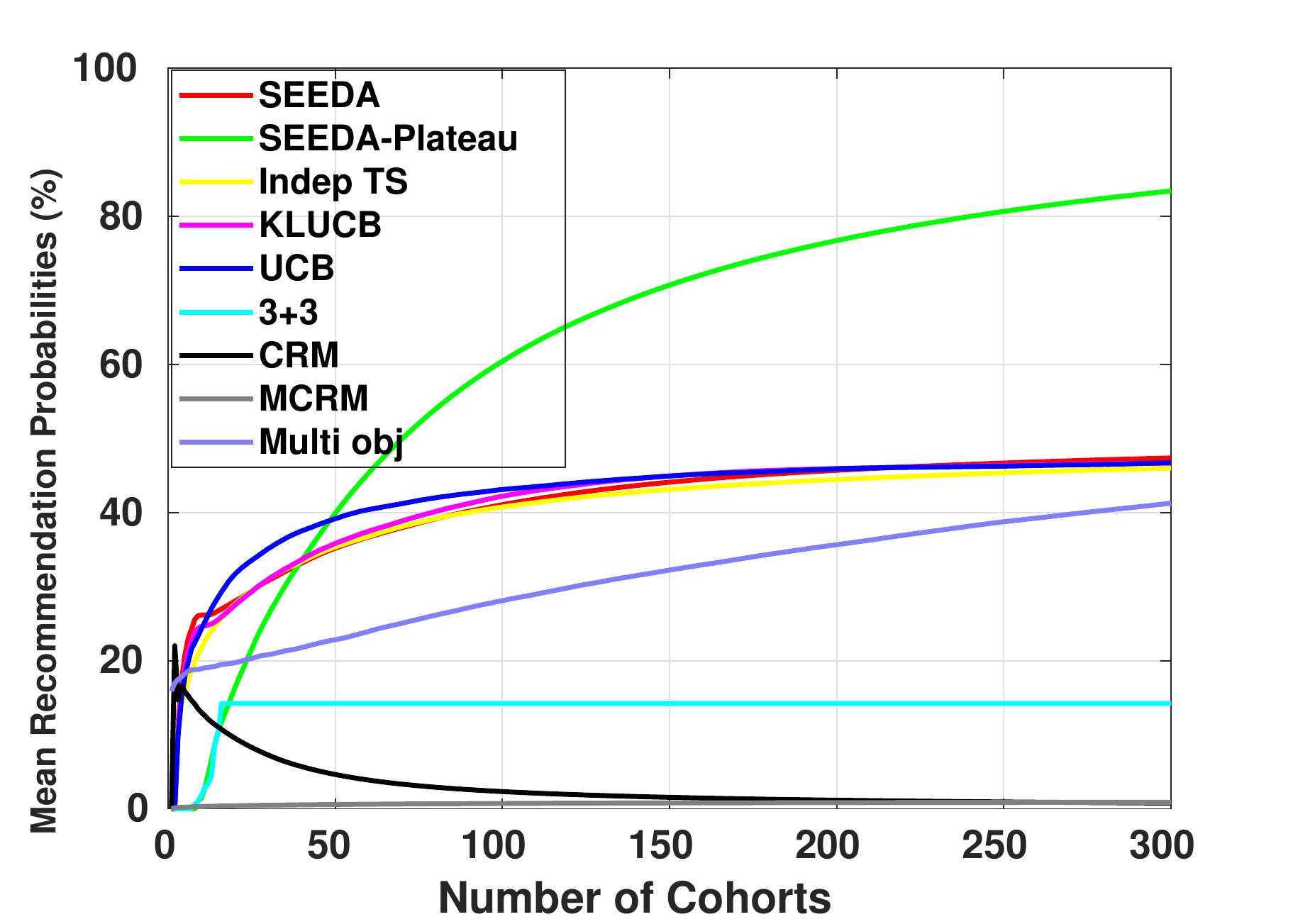}}  }
        }
    \caption{Mean allocation (left) and recommendation (right) probabilities versus number of patients $n$.}
    \label{fig:app2meanalloc}
\end{figure}

In particular, Table~\ref{tab:set2} only reports the recommendation and allocation percentages for a given $n=100$. It is of interest to see how these metrics change with $n$. We plot the mean allocation and recommendation probabilities as a function of $n$ in Fig.~\ref{fig:app2meanalloc}. It can be seen that SEEDA-Plateau outperforms all other methods across a large range of $n$.

\section{Experiment of a new setting and its comprehensive results}
\label{appx:sim:setting1}

In the main paper, a setting that has the efficacy reaching the maximal value (the optimal dose) before toxicity hits MTD threshold is used. A different setting can be considered when maximum efficacy dose exceeds the MTD threshold. The experiment results for this setting (called ``setting 2'') is reported in this section. Unless otherwise stated, the parameters are the same as in Section~\ref{sec:sim} of the main paper.

\begin{figure}[H]
\centerline{
        \subfigure{
        { \includegraphics[width=0.45\textwidth]{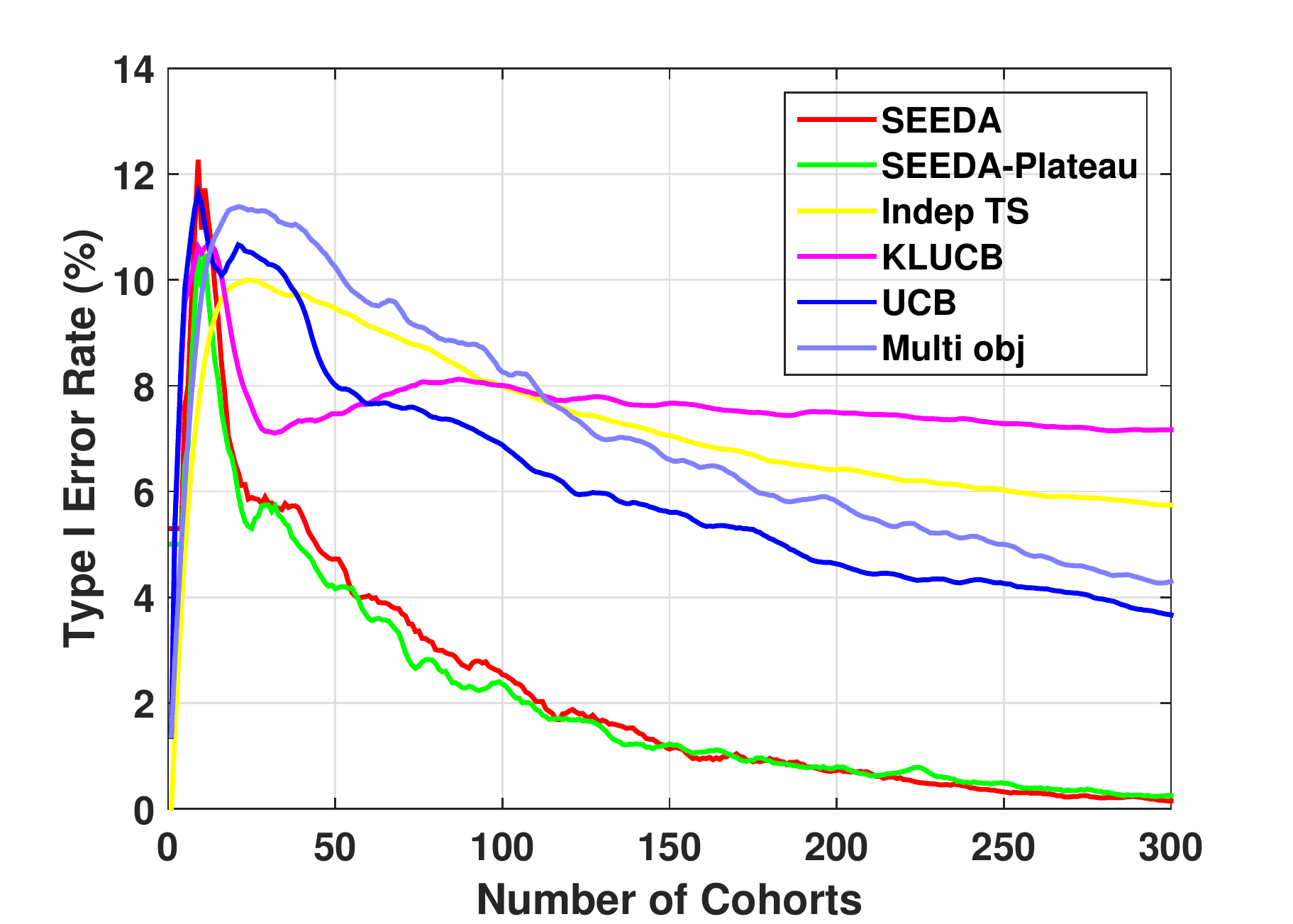}} }
        \hfil
        \subfigure{
        { \includegraphics[width=0.45\textwidth]{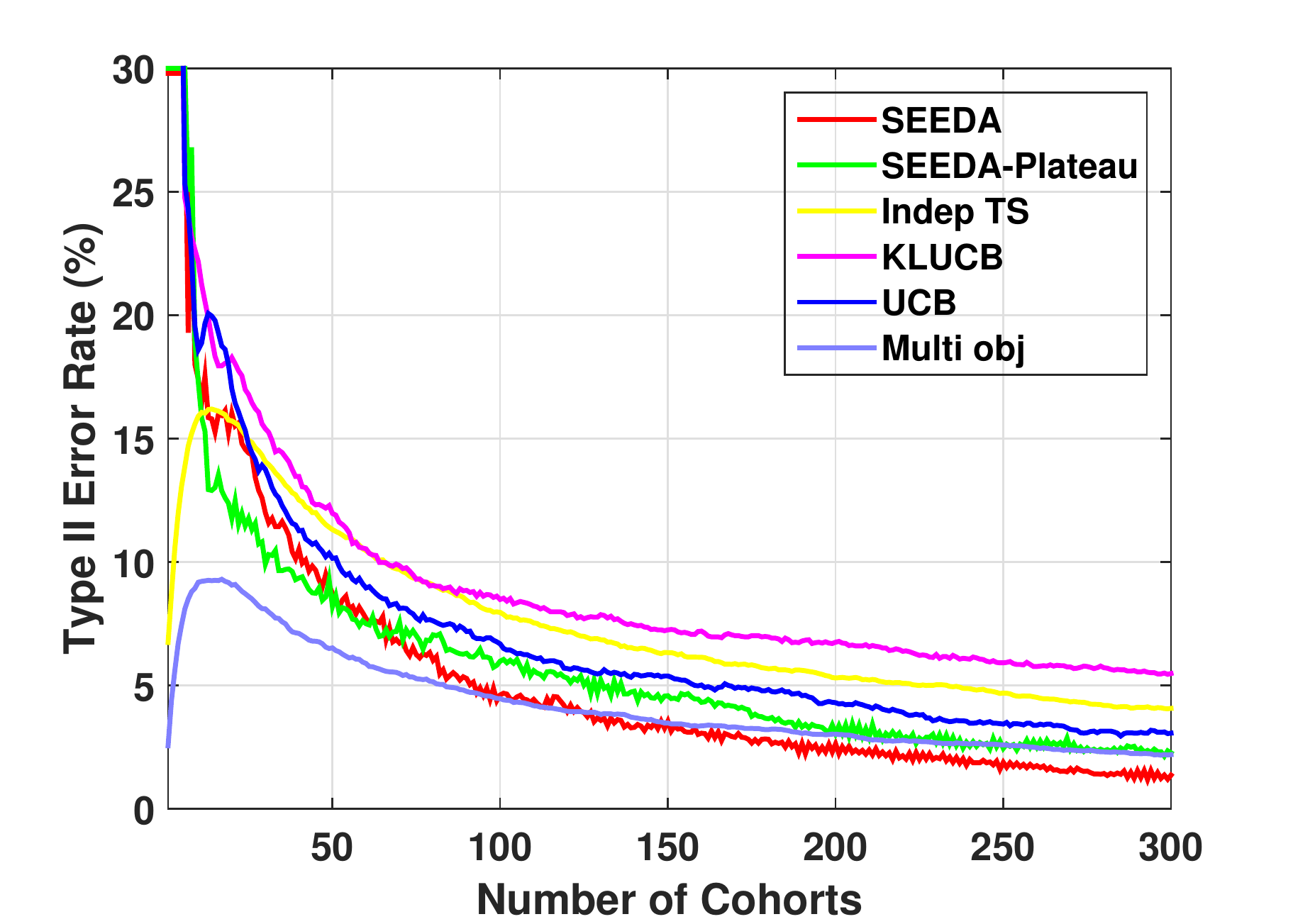}}  }
        }
    \caption{Type I and type II error rates in setting 2.}
    \label{fig:type-err-set1}
\end{figure}

Table~\ref{tab:set1} presents the setting as well as the allocation and recommendation percentages of each dose for all considered algorithms. For this scenario, dose level 3 is the optimal one. We note that a large portion of the previous conclusions in the main paper still hold. However, the gain of SEEDA-Plateau is less significant over SEEDA, but still outperforms all the comparing designs. The corresponding Type I and Type II error rates are similarly plotted in Fig.~\ref{fig:type-err-set1}.

\begin{table}[H]
\caption{Recommendation \& allocation percentages of different designs for setting 2.}
\label{tab:set1}
\begin{center}
\adjustbox{max width=\textwidth}{
\begin{tabular}{ c || c | c | c | c | c | c |||  c | c | c | c | c | c}
\hline
{} & \multicolumn{6}{c}{Recommended} & \multicolumn{6}{c}{Allocated} \\
\Xhline{3\arrayrulewidth}
Toxicity probabilities & 0.1& 0.2& \textbf{0.25} &0.4 &0.5 &0.6 &
 0.1& 0.2& \textbf{0.25} &0.4 &0.5 &0.6\\
\hline
Efficacy probabilities &0.3 &0.4& \textbf{0.5}& 0.7& 0.7& 0.7&
0.3 &0.4& \textbf{0.5}& 0.7& 0.7& 0.7\\ \Xhline{3\arrayrulewidth} 
\multirow{2}{*}{SEEDA} &   9.54 &	19.34&	52.66&	16.00&	2.12&	0& 6.82 &	17.61 &	48.99 &	21.77 &	3.47 &	1.33 \\&(3.40)  &(10.09) &(10.43) &(9.95)&(1.70)&(0)&(3.34)&(5.56)&(9.60)&(1.07)&(1.32)&(0.61)
 \\
\hline
\multirow{2}{*}{SEEDA-Plateau} &5.15&   34.51&   53.27&    5.84&    1.05&    0.01&
3.61 &   11.79 &   70.30 &   11.97 &    2.16 &    0.17 \\&(3.72)&(5.96)&(6.80)&(2.64)&(0.50)&(0)&(2.28)&(1.79)&(7.51)&(5.12)&(0.42)&(0.12)
\\
\hline
\multirow{2}{*}{Independent TS}& 22.61 &	22.12 &	29.05&	19.22&	4.50 &	2.50 & 2.58 &	3.17&	5.56 &	30.35 &	32.92 &	25.43 \\&(5.61) &(7.43)&(8.24) &(5.96) &(2.41)&(2.01)&(1.90)&(2.23 )&(3.72)&(4.73)&(4.62)&(3.82)
 \\
\hline
\multirow{2}{*}{KL-UCB}  &19.72  &	21.03 &	29.19 &	24.02 &	5.46&	0.59 & 2.13 &	2.50 &	3.37 &	32.80 &	30.63 & 28.58 \\&(3.65)&(4.14)&(9.27)&(5.44)&(1.88)&(0.38)&(0.48 )&(0.78)&(1.35)&(3.77 )&(8.16)&(6.99)

\\
\hline
\multirow{2}{*}{UCB}& 8.95 &	22.45&	41.04 &	21.61 &	4.83 &	1.11 & 8.12  &	10.31 &	13.20&	22.90&	22.58 &	22.89 \\
&(3.77)&(7.99)&(8.20)&(3.65)&(4.56)&(1.18)&(0.88)&(1.13)&(1.47) &(2.13) &(1.75)&(2.89)
\\
\hline
\multirow{2}{*}{3+3} &6.80 & 20&	23.80&	29.80 & 16.40& 3.20&
26.99 &	27.50& 19.59 & 13.14 & 5.01 & 0.76\\
&(0.12)&(13.40)&(10.24) &(8.45)&(5.45)&(3.12)&(2.89)&(3.25) &(1.45)&(0.25)&(1.25)&(0.75)
\\
\hline
\multirow{2}{*}{CRM} &0&	0&	0&	0&	99.10 &	0.90 &
0	&0&	0&	0&	99.11 & 0.89 \\&(0)&(0)&(0)&(0)&(0.42)&(0.36)
&(0)&(0)&(0)&(0)&(0.23)&(0.14)
\\
\hline
\multirow{2}{*}{MCRM} &0& 0.60 &	28.40 &	67.80 & 3.20 & 0&
0.60&	0.33& 29.17& 52.37 & 11.35 & 3.18\\
&(0) &(0.93)&(13.29) &(13.95) &(3.06) &(0) &(0.09) &(0.12) &(9.47) &(13.95) &(4.34) &(1.92) 
\\
\hline
\multirow{2}{*}{Multi-obj}& 6.57 &	13.38&	50.95&	22.71&	4.44&	1.95  &
20.17 & 14.78 &	19.05 &	20.29 &	14.57 &	11.17 \\&(2.64) 
&(8.12) &(9.92) &(6.95) &(1.27) &(0.55) &(5.32)  &(2.02)&(3.95)&(3.25)&(5.56)&(3.58)
\\
\hline

\end{tabular}
}
\end{center}
\end{table}

\begin{figure}[H]
\centerline{
        \subfigure{
        { \includegraphics[width=0.45\textwidth]{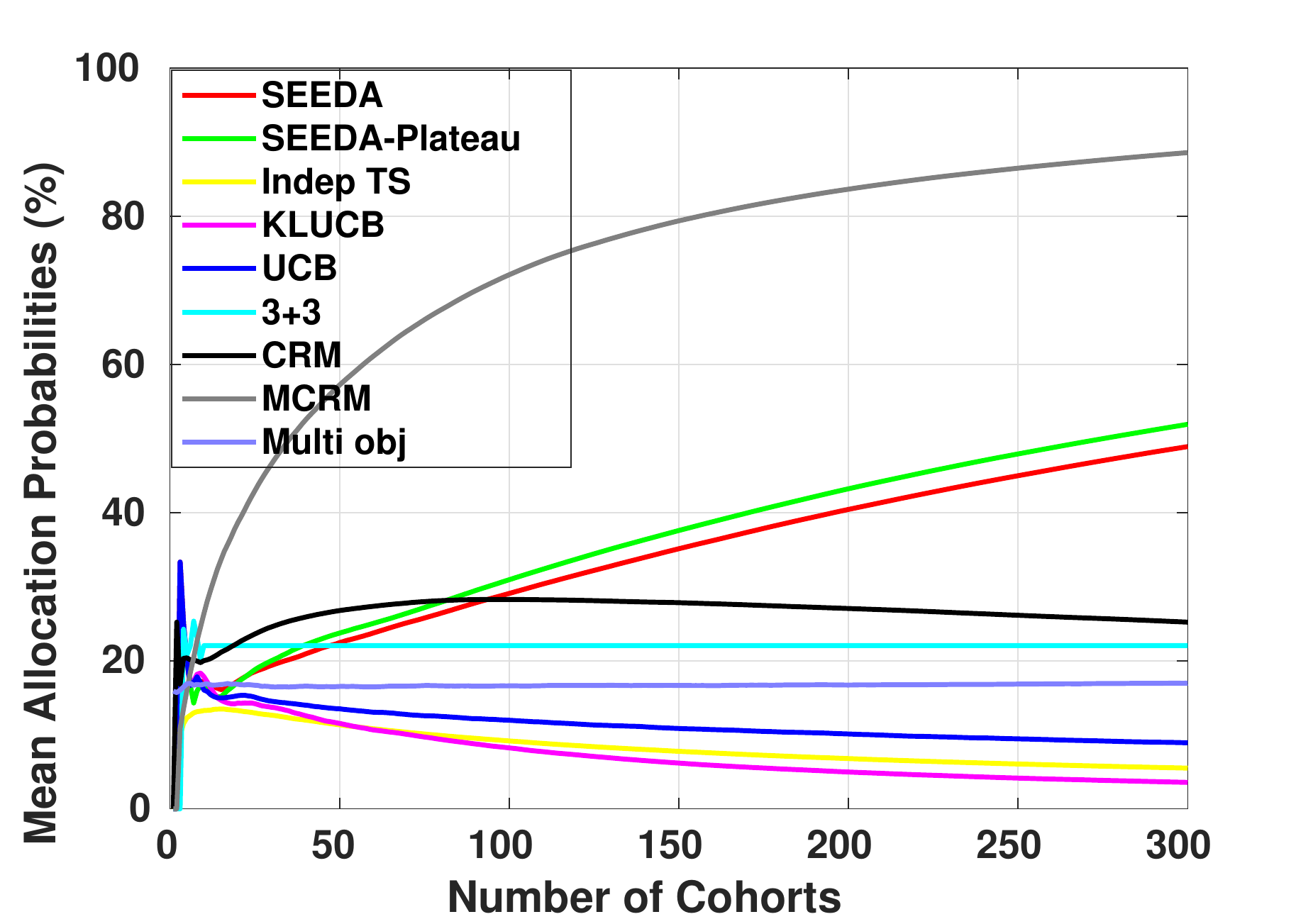}} }
        \hfil
        \subfigure{
        { \includegraphics[width=0.45\textwidth]{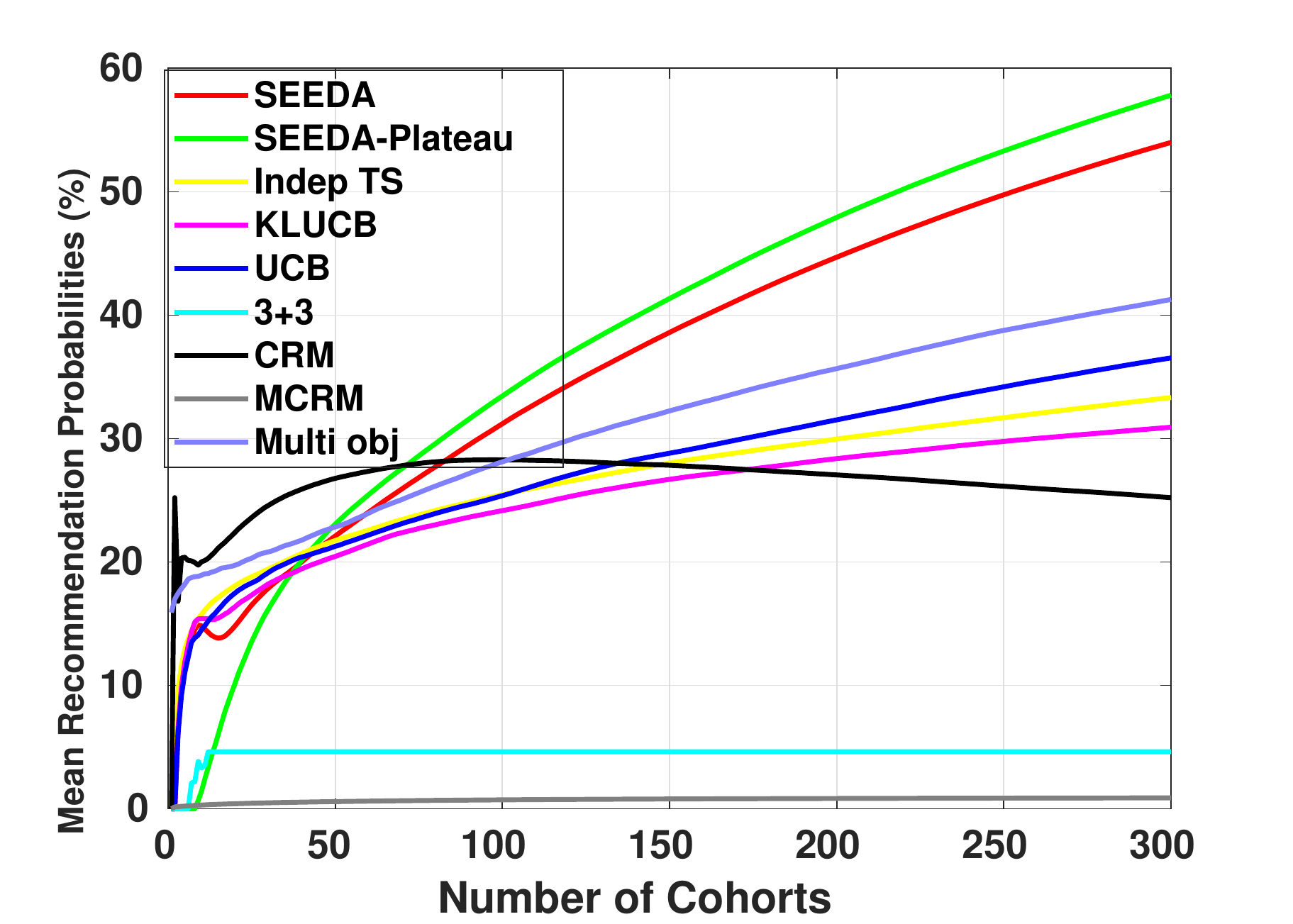}}  }
        }
    \caption{Mean allocation (left) and recommendation (right) probabilities versus number of patients $n$ in setting 2.}
    \label{fig:app2meanalloc2}
\end{figure}

An in-depth look at the mean allocation and recommendation probabilities versus number of patients $n$ for this new setting is given in Fig.~\ref{fig:app2meanalloc2}. The same observation as in Section~\ref{appx:sim:setting2} holds.

\begin{figure}[H] 
   \centerline{
        \subfigure {
       { \includegraphics[width=0.45\textwidth]{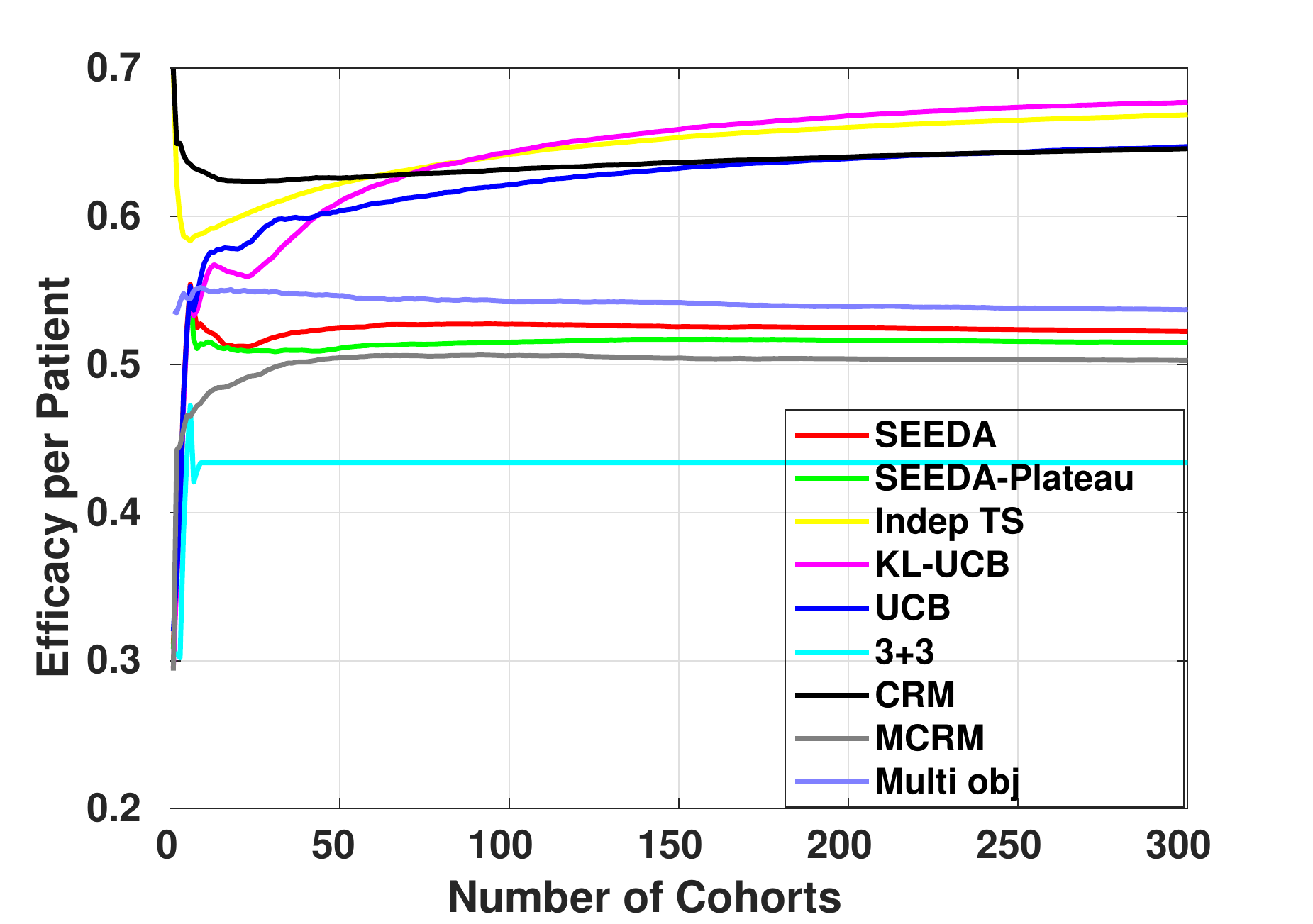}} }
        \hfil
        \subfigure {
        { \includegraphics[width=0.45\textwidth]{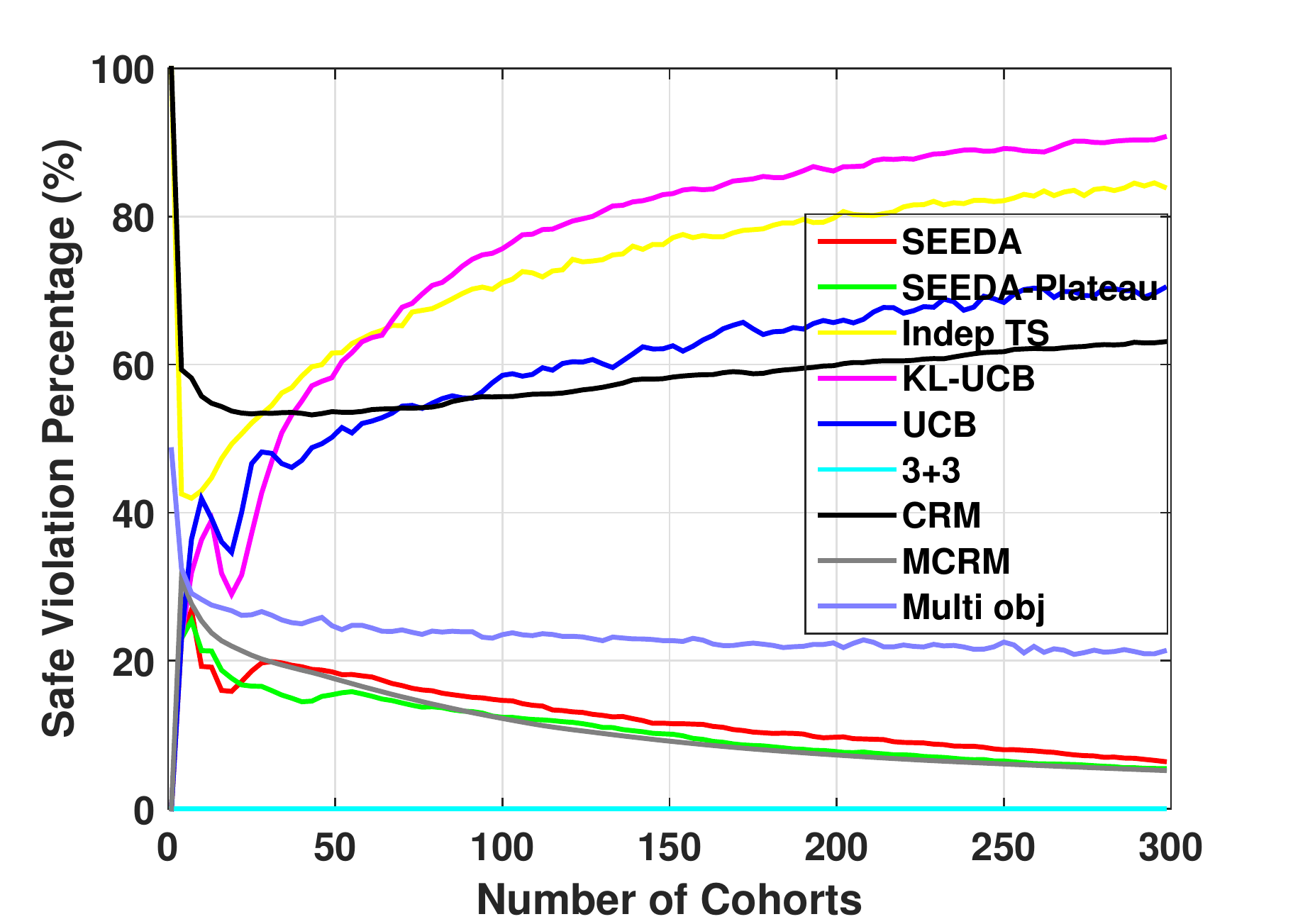}}  }
        }
    \caption{Comparison of efficacy per patient and the safety violation percentage in setting 2.}
    \label{fig:effvio1}
\end{figure}

The convergence of efficacy and toxicity as $t$ increases for setting 2 is plotted in Fig.~\ref{fig:effvio1}. There is a notable difference to the previous result in Fig.~\ref{fig:effvio2}, in that now SEEDA and SEEDA-Plateau converge to a different (but correct) dose than the other considered designs, which only emphasize maximum efficacy.  It is clear that with such aggressive pursue of efficacy, they succeed in obtaining better treatment effect than SEEDA(-Plateau), but at the significant cost of frequent violation of the safety constraint: as opposed to safety violation percentage hovering between $40\%$ and $50\%$ in Fig.~\ref{fig:effvio2}, now we face a violation in the range of $70\%$ to $90\%$ as shown in Fig.~\ref{fig:effvio1}.

\begin{figure}[H]
\centering
\centerline{\includegraphics[width=0.5\textwidth]{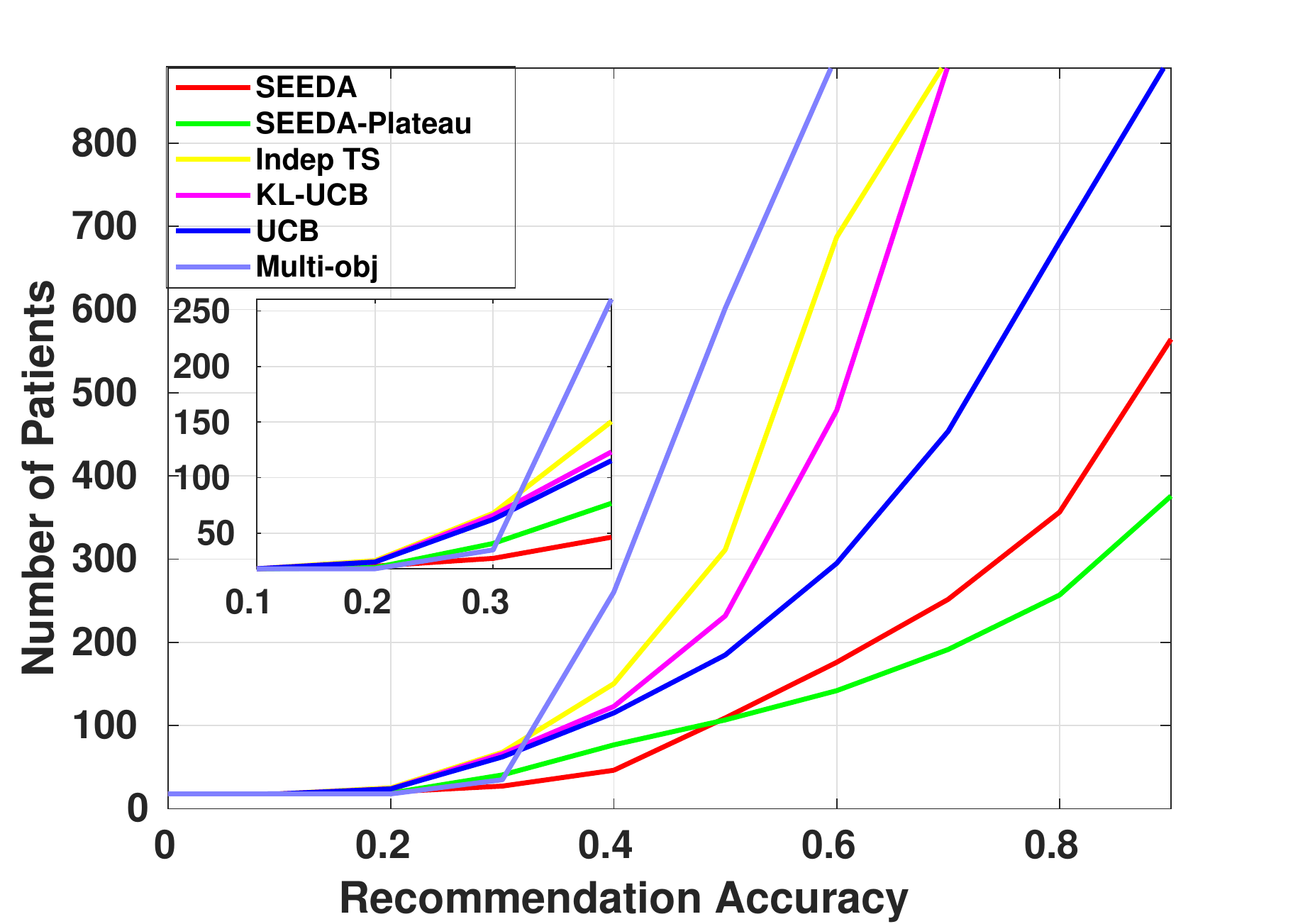}}
\caption{Sample size comparison in setting 2.}
\label{fig:earlystop1}
\end{figure}

Lastly, the sample efficiency is evaluated. Fig.~\ref{fig:earlystop1} plots the minimum number of patients to achieve a given a recommendation accuracy for different algorithms.

\section{Experiment setting 3 to 8 with evaluation of allocation and recommendation percentages}
\label{appx:sim:5setting}

This section reports the allocation and recommendation percentages of each dose for all considered algorithms under different toxicity/efficacy probabilities. We reuse the same 6 scenarios as those in the experiments of \cite{Zang2014}. See Table~\ref{table:appx:1} to \ref{table:appx:6} for the detailed results. They are in line with the conclusions of the main paper.

\begin{table}[h]
\caption{Recommended \& allocated percentages for Scenario 1 of \cite{Zang2014}.}
\label{table:appx:1}
\begin{center}
\adjustbox{max width=\textwidth}{
\begin{tabular}{ c || c | c | c |c | c |||  c | c |c | c | c}
\hline
{} & \multicolumn{5}{c}{Recommended} & \multicolumn{5}{c}{Allocated} \\
\Xhline{3\arrayrulewidth}
Toxicity probability & 0.08 & 0.12 & 0.2 & \textbf{0.3} & 0.4
& 0.08 & 0.12 & 0.2& \textbf{0.3}& 0.4\\
\hline
Efficacy probability &0.2& 0.4& 0.6& \textbf{0.8}& 0.55 &
0.2& 0.4& 0.6& \textbf{0.8}& 0.55\\
\Xhline{3\arrayrulewidth}
\multirow{2}{*}{SEEDA} &  2.72&    4.88&   21.72&   69.52&    1.16&
 2.84&	4.67&	18.55&	71.20&	2.74 \\&(1.01) &(2.14) &(7.50) &(10.11) &(0.62) &(0.78) &(1.95) &(6.04) &(7.65) &(2.74) 
 \\
\hline
\multirow{2}{*}{Indep TS} &2.34&   4.38&   12.91&   76.83&    3.54
&1.67&    2.99&   7.93&   81.18&    6.23 \\&(0.25) &(1.31) &(6.34) &(7.03) &(1.49)  &(0.62) &(0.64) &(0.36) &(2.55) &(2.44) 
\\
\hline
\multirow{2}{*}{KL-UCB}  &9.58&   23.99&  39.35&   24.27&    2.81&
3.24&   13.89&   30.91&   22.35&   29.61 \\
&(1.57) &(3.53) &(8.10) &(9.13) &(2.28) &(0.34) &(0.51) &(1.64) &(2.14) &(1.12) 
\\
\hline
\multirow{2}{*}{UCB}& 3.04&   12.41&   46.91&   35.24&    2.40&
10.91&   18.41&   33.34 &   28.32&   9.02\\
&(0.91) &(3.11) &(8.68) &(7.68) &(1.99)  &(0.72) &(1.31) &(2.10) &(2.67) &(1.85) 
\\
\hline
\multirow{2}{*}{3+3} & 4	&10.40&	20	&22.80&	42.80&
23.38&	22.81&	20.92&	15.80&	10.79\\
&(2.65) &(4.73) &(5.94) &(2.73) &(6.95)  &(5.79) &(1.22)&(4.63) &(2.14) &(1.26)  
\\
\hline
\multirow{2}{*}{CRM} & 0.09&  0.20&    1.72&   42.51&   55.48
& 0.09&    0.20&   1.72&   42.51&  55.48\\
&(0.02) &(0.02) &(0.02) &(2.38) &(2.38)  &(0.02) &(0.02) &(0.02) &(2.38) &(2.38) 
\\
\hline
\multirow{2}{*}{MCRM} & 1.09 &	2.26 &	26.69 &	65.68 &	4.28 &
2.09 &	2.26 &	26.50 &	64.88 &	4.28 \\
&(1.01) &(2.20) &(7.69) &(9.26) &(2.10) &(1.31) &(2.20) &(6.68) &(8.25) &(0.13) 
\\
\hline
\multirow{2}{*}{Multi-obj} & 1.41&    4.56&   22.69&   67.31&    4.03  &
18.42&   20.69&  22.51&   31.41&    6.97  \\
&(1.13) &(3.97) &(8.44) &(9.93) &(3.29) &(1.31) &(2.40) &(6.67) &(8.25) &(1.23) 
\\
\hline
\end{tabular}}
\end{center}
\end{table}

\begin{table}[h]
\caption{Recommended \& allocated percentages for Scenario 2 of \cite{Zang2014}.}
\begin{center}
\adjustbox{max width=\textwidth}{
\begin{tabular}{ c || c | c | c |c | c |||  c | c |c | c | c}
\hline
{} & \multicolumn{5}{c}{Recommended} & \multicolumn{5}{c}{Allocated} \\
\Xhline{3\arrayrulewidth}
Toxicity probability & 0.01 &\textbf{0.05}& 0.10& 0.15& 0.3&
0.01& \textbf{0.05}& 0.10& 0.15 &0.3\\
\hline
Efficacy probability &0.6 & \textbf{0.8} & 0.5& 0.4& 0.2 &
0.6& \textbf{0.8} & 0.5 &0.4 &0.2\\
\Xhline{3\arrayrulewidth}
\multirow{2}{*}{SEEDA} &   6.3&	91.23&	1.45&	0.53&	0.08 &
 5.56&	87.26&	2.95&	2.14&	2.09\\&(0.90) &(3.18) &(1.02) &(0.34) &(0.08) &(3.11)  &(3.94) &(2.09) &(1.43) &(0.63)  
 \\
\hline
\multirow{2}{*}{Indep TS}&5.31&	92.09&	1.47&	0.64&	0.48 &
 7.99&	83.18&	4.27&	2.91&	1.65 \\&(4.95) &(1.32) &(1.08) 
 &(0.56) &(0.16) &(2.55) &(5.34) &(4.34) &(2.30) &(1.05) 
 \\
\hline
\multirow{2}{*}{KL-UCB}  &9.68&	87.66&	1.91&	0.66&	0.09 &
7.01 &	81.93&	3.03&	2.31&	5.72 \\
&(2.73) &(2.98) &(1.20) &(0.44) &(0.04) &(1.57) &(1.94) &(0.82) &(0.51) &(0.31) 
\\
\hline
\multirow{2}{*}{UCB}&8.58&	89.80&	1.26&	0.34&	0.03 &
21.06&	46.31&	15.07&	11.16&	6.40 \\
&(3.98) &(4.18) &(1.24) &(0.24) &(0.03) &(2.20) &(2.69) &(1.68) &(1.28) &(0.73) 
\\
\hline
\multirow{2}{*}{3+3} & 0.20&	1.80&	5.40&	13.80&	78.80&
16.71&	18.81& 19.40&	19.88&	19.75\\
&(0) &(0.32) &(0.78) &(2.37) &(8.34) &(3.35) &(3.65) &(3.78) &(3.14) &(4.65) 
\\
\hline
\multirow{2}{*}{CRM} & 0 &	0&	0&	9.98&	90.02 &
0	&0&	0&	9.97&	90.03 \\
&(0) &(0) &(0) &(0.42) &(0.42) &(0) &(0) &(0) &(1.25) &(1.43) 
\\
\hline
\multirow{2}{*}{MCRM} & 0.08&	0.17&	1.15 &	13.47&	85.13&
1.08 &	0.17 &	1.15& 13.44&	84.16 \\&(0)  &(0.02) &(1.00) &(0.44) &(0.04)&(0.27) &(0.09) &(0.61) &(5.76) &(6.73)
\\
\hline
\multirow{2}{*}{Multi-obj} & 6.07&	90.85&	1.93&	0.92&	0.22 &
34.88&	51.74&	7.11& 4.34&	1.93 \\&(1.74) &(1.86) &(0.54) &(0.30) &(0.11) &(7.26) &(6.81) &(2.41) &(1.20) &(0.50)  
\\
\hline

\end{tabular}}
\end{center}
\end{table}

\begin{table}[h]
\caption{Recommended \& allocated percentages for Scenario 3 of \cite{Zang2014}.}
\begin{center}
\adjustbox{max width=\textwidth}{
\begin{tabular}{ c || c | c | c |c | c |||  c | c |c | c | c}
\hline
{} & \multicolumn{5}{c}{Recommended} & \multicolumn{5}{c}{Allocated} \\
\Xhline{3\arrayrulewidth}
Toxicity probability & 0.06& 0.08& 0.14 &\textbf{0.2} &0.3&
0.06 &0.08 &0.14& \textbf{0.2} &0.3\\
\hline
Efficacy probability &0.2 &0.4& 0.6& \textbf{0.8}& 0.55 &
0.2 &0.4& 0.6& \textbf{0.8}& 0.55\\
\Xhline{3\arrayrulewidth}
\multirow{2}{*}{SEEDA} &1.84&	1.97 &	6.15 &	88.12 &	1.58 &
2.27&	2.54&	6.27 &	85.46 &	3.46 \\
&(0.71) &(1.10) &(2.86) &(3.22) &(1.00)  &(0.71) &(1.11)  &(2.86) &(3.22)  &(0.99) 
\\
\hline
\multirow{2}{*}{Indep TS}&0.76 &	1.55  &	5.49 &	89.85 &	2.35 & 1.67&	2.98&	8.17&	81.28&	5.89 \\
 &(0.45) &(0.93) &(3.71) &(5.09) &(1.73) &(0.48)  &(1.33) &(4.48) &(4.96) &(1.79) 
 \\
\hline
\multirow{2}{*}{KL-UCB}  &2.64 &	7.29 &	28.47 & 57.18 &	4.43 &
2.62 &	6.58 &	26.85 &	55.07 &	8.87 \\
&(0.54) &(1.15) &(3.22) &(3.58) &(1.41) &(0.54) &(1.15) &(3.22) &(3.58) &(1.41) 
\\
\hline
\multirow{2}{*}{UCB}&1.71 &	3.57&	19.04 &	72.89 &	2.79 &
8.33&	13.17 &	22.75 &	44.71&	11.04 \\
&(0.48) &(1.33)  &(4.48) &(4.96) &(1.79)  &(0.48) &(1.33) &(4.48) &(4.96)  &(1.79) 
\\
\hline
\multirow{2}{*}{3+3} & 2.20 &	4.80&	10.60&	18.80&	63.60&
19.77&	20.08& 20.43&	18.67	& 15.29\\
&(1.93) &(2.10) &(3.22) &(3.92) &(9.33) &(3.54) &(5.93) &(5.12) &(3.95) &(3.45) 
\\
\hline
\multirow{2}{*}{CRM} & 0	&0&	0&	4.37 &	95.63 &
0	&0	&0	&4.37&	95.63\\
&(0) &(0) &(0) &(0.69) &(0.69) &(0) &(0) &(0) &(0.66) &(0.66)  
\\
\hline
\multirow{2}{*}{MCRM} & 0.60&	0.87 &	3.57 &	31.89 &	63.07  &
1.60 &	0.87 &	3.57 &	31.68 &	62.28  \\
&(0.54)  &(0.15) &(3.22) &(3.58) &(1.41) &(0.98) &(1.26) &(2.97) &(8.86) &(10.58) 
\\
\hline
\multirow{2}{*}{Multi-obj} & 0.78&	2.07 &	8.67&	84.99&	3.49  &
16.43&	20.56&	21.56&	34.45&	7.00\\
 &(0.20) &(0.45) &(1.97) &(2.40) &(1.02) &(0.20) &(0.45) &(1.97) &(2.41) & (1.02)  
\\
\hline
\end{tabular}}
\end{center}
\end{table}

\begin{table}[h]
\caption{Recommended \& allocated percentages for Scenario 4 of \cite{Zang2014}.}
\begin{center}
\adjustbox{max width=\textwidth}{
\begin{tabular}{ c || c | c | c |c | c |||  c | c |c | c | c}
\hline
{} & \multicolumn{5}{c}{Recommended} & \multicolumn{5}{c}{Allocated} \\
\Xhline{3\arrayrulewidth}
Toxicity probability & 0.05 &0.1& \textbf{0.25} &0.5& 0.6&
0.05 &0.1& \textbf{0.25}& 0.5& 0.6\\
\hline
Efficacy probability &0.2 &0.4& \textbf{0.6}& 0.8& 0.55 &
0.2 &0.4 &\textbf{0.6} &0.8 &0.55\\
\Xhline{3\arrayrulewidth}
\multirow{2}{*}{SEEDA} &  3.43 &	12.15&	79.72 &	4.37 &	0&
 3.40&	11.05&	79.44&	5.00&	1.12 \\
 &(1.26) &(3.69) &(4.25) &(1.90) &(0) &(1.24) &(3.48) &(4.28) &(1.75) &(0.45) 
 \\
\hline
\multirow{2}{*}{Indep TS} &11.53&	24.58 &	58.58&	2.66&	2.65&
1.68&	3.02 &	8.50&	81.01&	5.79 \\
&(9.17) &(10.80) &(12.42) &(1.53) &(3.42)  &(0.99)  &(2.39) &(5.40) &(16.00) &(6.50) 
\\
\hline
\multirow{2}{*}{KL-UCB}  &24.60&	37.78&	28.34&	6.91&	2.37 &
1.91&	2.43&	3.41&	51.61& 40.64 
\\
&(6.65) &(14.78) &(14.62) &(2.00) &(2.78) &(0.32) &(0.52) &(1.41) &(1.89) &(1.06) 
\\
\hline
\multirow{2}{*}{UCB} &4.87 &	32.53 &	60.34&	1.84&	0.42 &
14.29 &	26.93&	40.69&	9.15 &	8.94 \\
&(5.17) &(10.80) &(14.42) &(1.52) &(0.42) &(0.72) &(1.31) &(2.11) &(2.63) &(1.85) 
\\
\hline
\multirow{2}{*}{3+3} & 3&	6.20&	34.20&	40.40&	16.20&
22.57&	22.82& 26.70&	17.10&	4.29\\
&(1.46) &(4.64) &(6.85) &(7.10) &(4.16) &(7.69) &(6.98) &(7.89) &(6.79) &(0.68) 
\\
\hline
\multirow{2}{*}{CRM} & 0&	0&	0&	95.56&	4.44 &
0&	0&	0&	95.23&	4.77\\
&(0) &(0) &(0) &(0.14) &(0.14) &(0) &(0) &(0) &(2.12) &(2.12)
\\
\hline
\multirow{2}{*}{MCRM} & 0.84 &	3.77&	88.03 &	7.17 &	0.19  &
1.84 &	3.77 &	87.04 &	7.16 &	0.19  \\
&(0.83) &(1.73) &(3.92) &(3.58) &(0.01) &(0.83) &(1.73) &(3.91) &(3.57) &(0.01) 
\\
\hline
\multirow{2}{*}{Multi-obj} & 3.64&	19.66&	70.79&	4.80&	1.11&
19.93&	23.96 &	23.97 &	26.16& 5.98 \\
&(0.66) &(4.87) &(5.13) &(1.18) &(0.41)   &(6.11) &(4.93) &(4.69) &(4.17) &(2.25) 
\\
\hline
\end{tabular}}
\end{center}
\end{table}

\begin{table}[h]
\caption{Recommended \& allocated percentages for Scenario 5 of \cite{Zang2014}.}
\begin{center}
\adjustbox{max width=\textwidth}{
\begin{tabular}{ c || c | c | c |c | c |||  c | c |c | c | c}
\hline
{} & \multicolumn{5}{c}{Recommended} & \multicolumn{5}{c}{Allocated} \\
\Xhline{3\arrayrulewidth}
Toxicity probability & 0.1 &\textbf{0.2}& 0.4& 0.5& 0.6&
0.1& \textbf{0.2} &0.4 &0.5& 0.6\\
\hline
Efficacy probability &0.1 &\textbf{0.3}& 0.5& 0.5& 0.5&
0.1 &\textbf{0.3} &0.5 &0.5& 0.5\\
\Xhline{3\arrayrulewidth}
\multirow{2}{*}{SEEDA} & 7.20&	74.95&	15.01&	2.50&	0&
 6.86&	67.46&	21.21&	3.22&	1.26 \\
 &(1.10) &(4.42) &(4.84) &(1.46) &(0) &(0.96) &(3.49) &(4.11) &(1.58) &(0.61) 
 \\
\hline
\multirow{2}{*}{SEEDA-Plateau} &12.60& 82.20&	4.60& 0.60&	0&
19.50&	56.46&	15.49& 7.56& 1.00\\
&(2.12)  &(5.45) &(2.12) &(0.40) &(0) &(5.12) &(9.23) 
&(4.56) &(1.23) &(0.54) 
\\
\hline
\multirow{2}{*}{Indep TS}& 21.59&	50.75 &	21.15 &	4.73 &	1.78 &
 2.67&	6.34 &	29.19 &	30.59 &	31.22 \\
 &(7.05)  &(10.00) &(11.41) &(1.91) &(1.42) &(1.52) &(6.28) &(6.32) &(6.41)&(6.14) 
 \\
\hline
\multirow{2}{*}{KL-UCB}  &23.64&	40.01 &	21.58 &	10.92&	3.85&
3.80 &	2.24 &	23.69 &	40.19 &	30.10
\\
 &(4.52) &(10.18) &(10.82) &(2.16 ) &(0.81)  &(0.75) &(1.38) &(10.60) &(9.94) &(10.93)  
\\
\hline
\multirow{2}{*}{UCB}&13.71&	75.24 &	8.66 &	1.85 &	0.54 &
18.75&	36.38 &	16.49 &	14.23 &	14.14 \\
&(1.63 ) &(9.14) &(5.79) &(0.79) &(0.96) &(0.64)  &(4.19) &(2.57) &(2.63) &(2.55) 
\\
\hline
\multirow{2}{*}{3+3} & 7.40&	21.20&	42.60&	21.80&	7.00&
29.03&29.38&	23.97&	8.35&	1.76\\
&(1.42) &(12.30) &(6.42) &(3.06) &(4.12) &(0.79) &(3.32) &(2.14) &(1.15) &(0.42) 
\\
\hline
\multirow{2}{*}{CRM} & 0&	0&	0&	94.72&	5.28 &
0	&0&	0&	94.39&	5.61 \\
&(0) &(0) &(0) &(0.04) &(0.05) &(0) &(0) &(0) &(0.02) &(0.02) 
\\
\hline
\multirow{2}{*}{MCRM} & 2.86 &	62.72 & 33.03& 1.25 &	0.14  &
3.86&	62.02 &	32.73 &	1.25 &	0.14 \\
&(0.80) &(1.66)  &(4.13)  &(4.05) &(0) &(0.80)  &(1.66) &(4.11) &(0.42) &(0.02)  
\\
\hline
\multirow{2}{*}{Multi-obj} & 9.56 &	60.18 &	23.51 &	5.38&	1.38  &
23.42 &	25.22 &	22.55 &	16.27 &	12.54 \\
&(0.58) &(3.92) &(4.17) &(1.00)  &(0.39) &(6.89) &(5.30) &(5.28) &(6.61) &(5.79) 
\\
\hline

\end{tabular}}
\end{center}
\end{table}

\begin{table}[h]
\caption{Recommended \& allocated percentages for Scenario 6 of \cite{Zang2014}.}
\label{table:appx:6}
\begin{center}
\adjustbox{max width=\textwidth}{
\begin{tabular}{ c || c | c | c |c | c |||  c | c |c | c | c}
\hline
{} & \multicolumn{5}{c}{Recommended} & \multicolumn{5}{c}{Allocated} \\
\Xhline{3\arrayrulewidth}
Toxicity probability &0.01& 0.03 &0.05 &\textbf{0.1}& 0.2&
0.01& 0.03& 0.05& \textbf{0.1}& 0.2\\
\hline
Efficacy probability &0.1& 0.3 &0.45& \textbf{0.6}& 0.6&
0.1& 0.3& 0.45& \textbf{0.6} &0.6\\
\Xhline{3\arrayrulewidth}
\multirow{2}{*}{SEEDA} & 1.47&	1.79&	5.12&	48.97&	42.32 &
 3.59 &	2.93&	5.89&	45.65&	41.94 \\
 &(0.45) &(1.16) &(3.94) &(10.31) &(12.35)  &(0.56) &(1.55) &(3.19) &(6.51) &(6.62)  
 \\
\hline
\multirow{2}{*}{SEEDA-Plateau} &0	&0.20&	3	&96&	0.80 &
4.20&	5.64&	13.73&	40.22&	36.18\\
&(0) &(0.05) &(1.38) &(5.72) &(0.56) &(3.75) &(2.45) &(5.42) &(9.85) &(4.75)  
\\
\hline
\multirow{2}{*}{Indep TS}&0.42&	1.24 &	5.20 &	47.46 &	45.67 &
 13.71&	18.37 &	22.33&	28.10 &	17.48  \\
 &(0.31)  &(0.86) &(3.13) &(12.35) &(12.22) &(1.06)  &(3.55) &(5.87) &(8.80) &(8.57) 
 \\
\hline
\multirow{2}{*}{KL-UCB}  &1.96 &	2.55 &	9.57 &	54.30&	31.62&
3.78 &	3.32 &	9.42 &	52.03 &	31.45
\\
&(0.50) &(1.46) &(3.46) &(10.30)  &(10.06)  &(0.77) &(0.76) &(2.14) &(10.56)  &(10.53) 
\\
\hline
\multirow{2}{*}{UCB}&1.31&	2.06&	9.47 &	56.47 &	30.69 &
8.18&	12.58 &	19.54 &	32.85 &	26.84  \\
&(0.37) &(1.22) &(4.06) &(10.82) &(10.74) &(0.58)  &(1.30) &(2.02) &(2.83) &(2.93) 
\\
\hline
\multirow{2}{*}{3+3} & 0&	1.40&	2.20&	8.20&	88.20 &
17.14&	18.15&	18.32&	20.07&	20.74\\
&(0) &(0.23) &(1.23) &(1.27) &(7.21) &(6.79) &(7.90) &(7.45) &(6.52) &(6.48)  
\\
\hline
\multirow{2}{*}{CRM} & 0&	0&	0	&65.39 &	34.61 &
0&	0&	0&	65.15&	34.85 \\
&(0) &(0) &(0) &(2.29) &(2.29) &(0) &(0) &(0) &(6.79) &(6.41) 
\\
\hline
\multirow{2}{*}{MCRM} & 0.06& 0.08 &	0.49 &	2.92 &	96.45 &
1.06 &	0.08 &	0.48&	2.92 &	95.45  \\
&(0.02)  &(0.04) &(0.50) &(1.21) &(3.00) &(0.25) &(0.04)  &(0.29) &(2.01) &(3.00)  
\\
\hline
\multirow{2}{*}{Multi-obj} &0.63&	1.60&	6.78&	49.01&	41.98  &
13.71 &	18.37&	22.33&	28.10&	17.48 \\
&(0.17) &(0.36) &(1.52) &(10.01) &(10.03) &(7.55) &(8.18) &(8.00) &(7.23) &(7.41) 
\\

\hline

\end{tabular}}
\end{center}
\end{table}

\end{document}